\documentclass[10pt,twocolumn,a4paper]{article}

\usepackage{cvpr}
\usepackage{times}
\usepackage{epsfig}
\usepackage{graphicx}
\graphicspath{{supplement/}}
\usepackage{amsmath}
\usepackage{amssymb}
\usepackage{tabu}
\usepackage{booktabs}
\usepackage{caption}
\usepackage{subcaption}
\usepackage{tikz}
\usepackage[hyphens,spaces,obeyspaces]{url}

\usepackage{algpseudocode}
\usepackage{algorithm}

\usepackage[inline]{enumitem}
\usepackage{siunitx}
\usepackage{placeins}
\usepackage{ifthen}
\newboolean{my_hasappendix_check}

\renewcommand{\textfraction}{0.0}

\usepackage[pagebackref=true,breaklinks=true,letterpaper=true,colorlinks,bookmarks=false,citecolor=blue]{hyperref}

\cvprfinalcopy
\def\cvprPaperID{0000}

  \newcommand*\samethanks[1][\value{footnote}]{\footnotemark[#1]}

\DeclareRobustCommand{\appendixcheck}[3]{%
  \ifthenelse{\boolean{#1}}{#2}{#3}
}

\begin{document}
\setboolean{my_hasappendix_check}{true}

\title{PatchPerPix for Instance Segmentation}

\author{%
  Peter Hirsch\thanks{equal contribution, listed in random order}\hspace{2em}%
  Lisa Mais\samethanks\hspace{2em}%
  Dagmar Kainmueller\\%
{\normalsize Max Delbrück Center for Molecular Medicine in the Helmholtz Association}\\
{\tt\small \{firstname.lastname\}@mdc-berlin.de}
}

\maketitle

\begin{abstract}
  We present a novel method for proposal free instance segmentation that can handle sophisticated object shapes which span large parts of an image and form dense object clusters with crossovers. 
  Our method is based on predicting dense local shape descriptors, which we assemble to form instances. All instances are assembled simultaneously in one go. 
  To our knowledge, our method is the first non-iterative method that yields instances that are composed of learnt shape patches. 
We evaluate our method on a diverse range of data domains, where it defines the new state of the art on four benchmarks, namely the ISBI 2012 EM segmentation benchmark, the BBBC010 C.\,elegans dataset, and 2d as well as 3d fluorescence microscopy data of cell nuclei. We show furthermore that our method also applies to 3d light microscopy data of Drosophila neurons, which exhibit extreme cases of complex shape clusters. \\Code: \url{https://github.com/Kainmueller-Lab/PatchPerPix}

\end{abstract}

\vspace{-0.05em}

\section{Introduction}
The task of instance segmentation has a wide range of applications in natural images as well as microscopy images from the biomedical domain. 
A prevalent class of instance segmentation methods, namely proposal-based methods based on RCNN~\cite{rcnn_girshick2014,maskrcnn_he2017}, has proven successful in cases where instance location and size can be well-approximated by bounding boxes.
However, in many cases, especially in the biomedical domain, this does not hold: Instances may span widely across the image, and hence multiple instances may have very similar, large bounding boxes. To complicate things, instances may be densely clustered, in some cases overlapping, including crossovers. 
Proposal-free methods are applicable in such cases, where popular choices include metric learning / instance coloring~\cite{de2017semantic,dense_embeddings_lee2019,chen2019instance,Kulikov_2020_CVPR}, affinity-based methods \cite{connectome_funke2018,mutex_wolf2018,graph_merge_liu2018,gao2019ssap}, and learnt watershed~\cite{deepws_bai2016,wolf2017learned}. However, respective pixel-wise predictions do not explicitly capture instance shape, nor are they suitable for disentangling overlapping instances. 

To overcome these limitations, we propose to (1)~densely predict representations of the shapes of instance patches, (2)~cover the image foreground with the most plausible shape patches, and (3) puzzle together complete instance shapes from these patches by means of partitioning a patch affinity graph. The approach of covering the image by selecting from a redundant set of instance patch predictions allows for naturally handling overlap (including crossovers), as overlapping instance patches can be selected, potentially resulting in pixels covered by multiple instances. 

Our general idea is closely related to Singling Out Networks~\cite{singlingout_yurchenko2017}. However, they are different in that they rely on a dictionary of known instances, thereby limiting the variability of objects they can handle, and they only consider predicting whole instances and not patches of instances, thereby limiting the size of feasible object categories. 

\begin{figure*}[tb]
    \centering
	\def\svgwidth{\textwidth}
	{\small\input{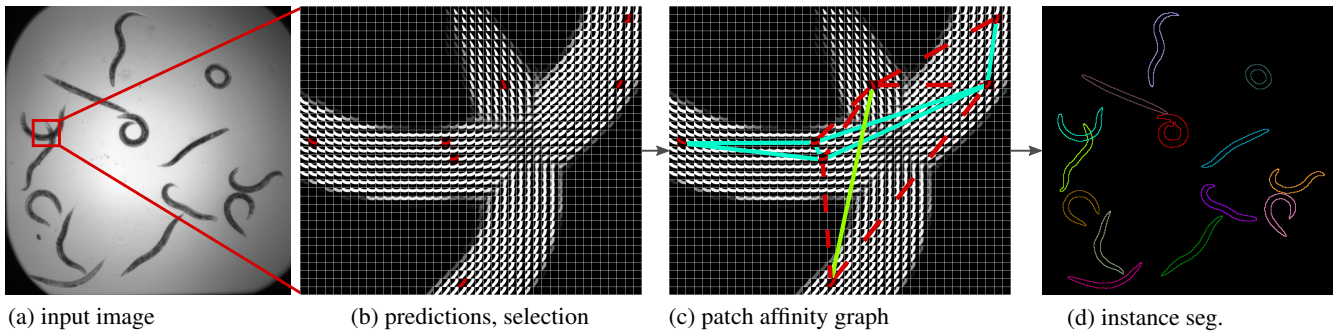}}
	\caption{PatchPerPix overview. Given the raw input image (a), a CNN predicts dense patches for each pixel (b, best seen with zoom) which are then used to find a consensus for each pair of pixels within the patch size. The patches that best agree with this consensus are selected (shown in red in b) and connected to form a patch affinity graph. (c) Edges of the patch affinity graph are assigned scores  derived from the agreement of the \emph{merged} shape patches with the consensus. The final instance segmentation (d) is obtained by signed graph partitioning. Shown in (c,d) is the result of connected component analysis on the positive subgraph, where edges with negative scores are depicted in red. }	
    \label{fig:pipeline}
\end{figure*}

Our shape prediction network predicts, for each pixel of the input image, a representation of the local shape of the instance this pixel belongs to, namely a \textit{shape patch} of the pixel's instance. The architecture we propose is derived from the U-Net~\cite{unet_ronneberger2015}, thus allowing for efficient dense prediction. 
As representations of instance patch shapes, we explore local binary masks, as well as encodings (i.e.\ compressed versions) of these. The idea of predicting instance shape masks per pixel of an image has been pursued before~\cite{tensormask_chen2019,instanceFCN_dai,encoding_shapes_jetley2016}. However, all these approaches work on the assumption that a shape mask can capture a complete instance shape. Thus they are designed for object categories common to natural images rather than for disentangling clusters of complex shapes that occupy similar bounding boxes, as relevant in the biomedical domain. 
Predicting shape encodings instead of binary masks is also not new~\cite{encoding_shapes_jetley2016}. However, 
besides only considering complete instance shapes as opposed to our patches of instances, in~\cite{encoding_shapes_jetley2016},  shape encoding and respective decoder are trained separately, where we show in our work that end-to-end training yields considerable improvement. 

The variant of our method that predicts local binary masks as shape representations is closely related to methods that employ long-range affinities~\cite{keuper2015efficient,mutex_wolf2018,graph_merge_liu2018,gao2019ssap}. In essence, our predicted binary patches can be interpreted as dense affinities in a neighborhood around each pixel. However, in contrast to affinity-based methods, we instead interpret our predictions as patches of instances, from which we puzzle together complete instances. This way, our yielded global instance shapes are assembled from \emph{learned shape patches}, a property that does not hold for affinity-based methods. 
Note that in this respect, our method is related to CELIS~\cite{celis_maitin2016}, which learns to agglomerate super-pixels to form instances with plausible shapes, yet their initial pixel-wise predictions do not capture object shape. Furthermore, our method is related to Flood Filling Networks~\cite{ffn_2016}, an iterative method that learns to expand instances one-by-one. In contrast, our method segments all instances simultaneously in one pass. 

We show in a quantitative evaluation that our method is the new state of the art on the ISBI 2012 challenge on segmentation of neuronal structures in EM stacks~\cite{arganda2015crowdsourcing}, outperforms the previous state of the art~\cite{singlingout_yurchenko2017,semiconv_novotny2018,Kulikov_2020_CVPR,chen_2022_layering} on the BBBC010 benchmark dataset of worm images~\cite{celegans_ljosa2012} by a large margin, and also outperforms the state of the art~\cite{stardist_schmidt2018,stardist3d_weigert2019,hirsch20_an_auxil_task_for_learn_nucle_segme_in_3d_micro_image} on 2d and 3d light microscopy images of densely packed cell nuclei. Last but not least, we demonstrate that our method also applies to the complex tree-like shapes of neurons in 3d light microscopy images.

In summary, our contributions are:
\begin{itemize}
\item A novel method for segmenting instances of complex shapes that spread widely across an image in crowded scenarios, with overlaps and crossovers. 
\item Instance segmentations are assembled from learnt shape pieces. Our method is, to our knowledge, the first such method that is not iterative, i.e.\ we compute all instances in one pass.
\item Our method defines the new state of the art on the competitive ISBI 2012 EM segmentation challenge, considerably outperforms the state of the art on the challenging BBBC010 C.\,elegans dataset, and also defines the new state of the art  on 2d and 3d benchmark data of cell nuclei.
\end{itemize}


\section{PatchPerPix for Instance Segmentation}
We train a CNN to predict dense local shape patches, from which we assemble all instances in an image simultaneously in a one-pass pipeline. Figure~\ref{fig:pipeline} and \ifthenelse{\boolean{my_hasappendix_check}}{Suppl.\ Fig.~\ref{fig:pipeline_detail_suppl}}{Suppl.\ Fig.~5} provide an overview of our proposed method, which we term \textit{PatchPerPix}.

Formally, our CNN yields an estimate $p \colon \textnormal{Dom}(I) \times \mathcal{P} \to \left[0,1\right]$ of the function
\begin{align*}
p^{*} \colon & \textnormal{Dom}(I) \times \mathcal{P} \to \left\{0,1\right\} \\
  & (\mathbf{x},\mathbf{dx}) \mapsto 
  \begin{cases}
    1 & \ \textnormal{if} \ \textnormal{Instance}(\mathbf{x}) = \textnormal{Instance}(\mathbf{x}+\mathbf{dx})\\
      & \ \textnormal{and} \ \mathbf{x}, \mathbf{x}+\mathbf{dx} \in \textnormal{fg}(I) \\
	 0 &\ \textnormal{otherwise}
	\end{cases}
\end{align*}
that captures, for each pixel $\mathbf{x} \in \mathcal{R}^d$ in the foreground $\textnormal{fg}(I)$ of a $d$-dimensional image $I$,
and each pixel $\mathbf{x}+\mathbf{dx}$ at a fixed, dense set of offsets $\mathcal{P} \subset \mathcal{R}^d$,
whether $\mathbf{x}$ and $\mathbf{x}+\mathbf{dx}$ belong to the same instance.

Section~\ref{subsec:instance_assembly} describes our proposed instance assembly pipeline given the estimated function~$p$. Section~\ref{subsec:architecture} describes the CNN architectures we explore to yield $p$.

\subsection{Instance Assembly}
\label{subsec:instance_assembly}
We denote a restriction of the estimated function $p$ to a single pixel as
\begin{align*}
p_\mathbf{x} \colon \mathbf{x}+\mathcal{P} \to \left[0,1\right], \
\mathbf{y} \mapsto p(\mathbf{x},\mathbf{y-x})
\end{align*}
We denote the domain of $p_{\mathbf{x}}$ as $\textnormal{patch}(p_{\mathbf{x}}) := \mathbf{x}+\mathcal{P}$.
For each patch, the pixels that are predicted to belong to the same instance as $\mathbf{x}$ by means of a probability threshold $t$, i.e.\ the pixels classified as foreground w.r.t.\ the instance at $\mathbf{x}$, are denoted as
\begin{equation}
\textnormal{fg}(p_\mathbf{x}) := \left\{ \mathbf{y} \in \textnormal{patch}(p_\mathbf{x}) : p_\mathbf{x}(\mathbf{y})>t \right\},
\label{eq:patch_fg}
\end{equation}
and, accordingly, the respective background pixels as
\begin{equation}
\textnormal{bg}(p_\mathbf{x}) := \left\{ \mathbf{y} \in \textnormal{patch}(p_\mathbf{x}) : p_\mathbf{x}(\mathbf{y})<1-t \right\}.
\end{equation}
For each pixel pair $(\mathbf{y},\mathbf{z})$ covered by at least one informative patch, i.e.\ $\exists \mathbf{x} \in \textnormal{Dom}(I): \{\mathbf{y}, \mathbf{z} \}\subset \textnormal{patch}(p_\mathbf{x}) \wedge  \{\mathbf{y}, \mathbf{z}\} \cap \textnormal{fg}(p_\mathbf{x}) \neq \emptyset $, summing up observations from all patches yields a consensus that $\mathbf{y}$ and $\mathbf{z}$ belong to the same instance, i.e.\ a consensus affinity
\begin{equation}
\label{eq:aff}
\begin{aligned}
\textnormal{aff}&(\mathbf{y},\mathbf{z}) :=  \frac{1}{Z_{\textnormal{aff}}(\mathbf{y},\mathbf{z})} \cdot 
 \Big( \sum_{\substack{\mathbf{x}\in\textnormal{Dom}(I): \\ \{\mathbf{y},\mathbf{z}\} \subset \textnormal{fg}(p_\mathbf{x})}} p_{\mathbf{x}}(\mathbf{y})\cdot p_{\mathbf{x}}(\mathbf{z}) \\
 & -  \sum_{\substack{\mathbf{x}\in\textnormal{Dom}(I): \\ \mathbf{y} \in \textnormal{fg}(p_\mathbf{x}), \mathbf{z} \in \textnormal{bg}(p_\mathbf{x}) }} p_{\mathbf{x}}(\mathbf{y})\cdot (1-p_{\mathbf{x}}(\mathbf{z}) ) \\
 & -  \sum_{\substack{\mathbf{x}\in\textnormal{Dom}(I): \\ \mathbf{y} \in \textnormal{bg}(p_\mathbf{x}), \mathbf{z} \in \textnormal{fg}(p_\mathbf{x})}} (1-p_{\mathbf{x}}(\mathbf{y}))\cdot p_{\mathbf{x}}(\mathbf{z}) \ \Big)
\end{aligned}
\end{equation}
with normalization factor 
\begin{equation}
  \begin{aligned}  Z_{\textnormal{aff}}&(\mathbf{y},\mathbf{z}) := | \{ \mathbf{x} \in \textnormal{Dom}(I): \\
    &
 \{\mathbf{y}, \mathbf{z} \}\subset \textnormal{patch}(p_\mathbf{x}) \wedge  \{\mathbf{y}, \mathbf{z}\} \cap \textnormal{fg}(p_\mathbf{x}) \neq \emptyset \} |.
\end{aligned}
\end{equation}
Given these consensus affinities, we define a score
\begin{equation}
\label{eq:score}
\begin{aligned}
\textnormal{sc}&\textnormal{ore}(p_\mathbf{x}) := \frac{1}{ Z_{\textnormal{score}}(p_\mathbf{x}) } \cdot 
\Big( \\
 & \sum_{ \{\mathbf{y},\mathbf{z}\} \subset \textnormal{fg}(p_\mathbf{x})}  \textnormal{aff}(\mathbf{y},\mathbf{z}) -\sum_{\substack{\mathbf{y}\in \textnormal{fg}(p_\mathbf{x}), \\ \mathbf{z} \in \textnormal{bg}(p_\mathbf{x})}} \textnormal{aff}(\mathbf{y},\mathbf{z})   \ \Big)
\end{aligned}
\end{equation}
with normalization factor
\begin{equation}
\begin{aligned}
  Z_{\textnormal{score}}&(p_\mathbf{x}) :=\vert \{ \{\mathbf{y},\mathbf{z}\} \subset \textnormal{patch}(p_\mathbf{x}): \\
  & \{\mathbf{y}, \mathbf{z}\} \cap \textnormal{fg}(p_\mathbf{x}) \neq \emptyset \}\vert
\end{aligned}
\end{equation}
for each patch with non-empty foreground by assessing how well it agrees with the consensus.
We rank all patches w.r.t.\ their score (Eq.~\ref{eq:score}).
We employ a greedy set cover algorithm to select high-ranking patches whose patch foregrounds $\textnormal{fg}(p_{\mathbf{x}})$ fully cover the image foreground $\textnormal{fg}(I)$.
Section~\ref{subsec:architecture} describes how we obtain the image foreground.
In more detail, the set cover algorithm proceeds as follows: Iterating from high to low score over the ranked list of patches, we pre-select patches if they cover previously uncovered image foreground, until the image foreground is fully covered. We further thin out this pre-selection as follows: We iteratively select as next patch from the pre-selection the patch that covers the most remaining foreground, until the whole foreground is covered.

Given this selection of high-ranking patches, the consensus affinities (Eq.~\ref{eq:aff}) allow us to define a score that measures for a \emph{pair of patches} whether they belong to the same instance, i.e.\ a consensus affinity between $p_\mathbf{x}$ and $p_\mathbf{y}$
\begin{equation}
\label{eq:paff}
\begin{aligned}
\textnormal{paff}&(p_\mathbf{x},p_\mathbf{y}) := 
 \frac{1}{ Z_{\textnormal{paff}}(p_\mathbf{x},p_\mathbf{y}) } \cdot \sum_{\substack{\mathbf{v}\in \textnormal{fg}(p_\mathbf{x}), \\ \mathbf{w} \in \textnormal{fg}(p_\mathbf{y})}}\textnormal{aff}( \mathbf{v}, \mathbf{w} ) 
 \end{aligned}
\end{equation}
with normalization factor 
\begin{equation}
\begin{aligned} Z_{\textnormal{paff}}&(p_\mathbf{x},p_\mathbf{y}) := \vert\{\mathbf{v}\in\textnormal{fg}(p_\mathbf{x}), \mathbf{w}\in\textnormal{fg}(p_\mathbf{y}): \\
	&\exists \mathbf{z} :\{ \mathbf{v} ,\mathbf{w} \} \subset \textnormal{patch}(p_\mathbf{z}) \wedge \{ \mathbf{v} ,\mathbf{w} \} \cap \textnormal{fg}(p_\mathbf{z}) \neq \emptyset
	\}\vert .
\end{aligned}
\end{equation}

We compute patch pair affinities (Eq.\ \ref{eq:paff}) between selected high-ranking patches iff the respective $Z_\textnormal{paff}(\cdot,\cdot)>0$, yielding a patch affinity graph. We partition this graph via connected component analysis on the positive subgraph, or alternatively by means of the mutex watershed algorithm~\cite{mutex_wolf2018}, depending on the application domain.
We obtain the final instance segmentation by assigning, per connected component, a unique instance ID to all pixels contained in the union of the respective patch foregrounds.
Note that in general, this may assign multiple instance IDs to some pixels, which is desired in some, but not all, applications. In case overlapping instances are not desired, we assign the ID of the patch prediction with highest probability at the respective pixel.

We implemented the computationally expensive parts of our instance assembly pipeline in CUDA for efficient execution. In applications with sparse image foreground, we further improve computational efficiency by restricting $\textnormal{patch}(p_\mathbf{x})$ to the image foreground, i.e.\ $\textnormal{patch}_{\textnormal{sparse}}(p_\mathbf{x}) := \textnormal{patch}(p_\mathbf{x}) \cap \textnormal{fg}(I) .$

\subsection{CNN Architecture}
\label{subsec:architecture}

We train a deep convolutional neural network to predict the function $p$.
It does so by predicting $p_{\mathbf{x}}(\mathbf{x}+\mathcal{P})$ for each pixel of the input image.
Thus the cardinality of the set $\mathcal{P}$ determines the number of output channels of the network.
We train the network w.r.t.\ standard cross-entropy loss averaged over all outputs.
We use a U-Net \cite{unet_ronneberger2015} as backbone architecture.
To facilitate predictions of shape representations with hundreds of dimensions, we keep the number of feature maps fixed (instead of reducing) in the upward path of the U-Net. Thus we avoid having to predict high-dimensional pixel-wise outputs from only tens of feature maps as present in the penultimate layer of a standard \mbox{U-Net}.
\begin{figure*}[htbp]
  \centering
  \includegraphics[width=0.75\linewidth]{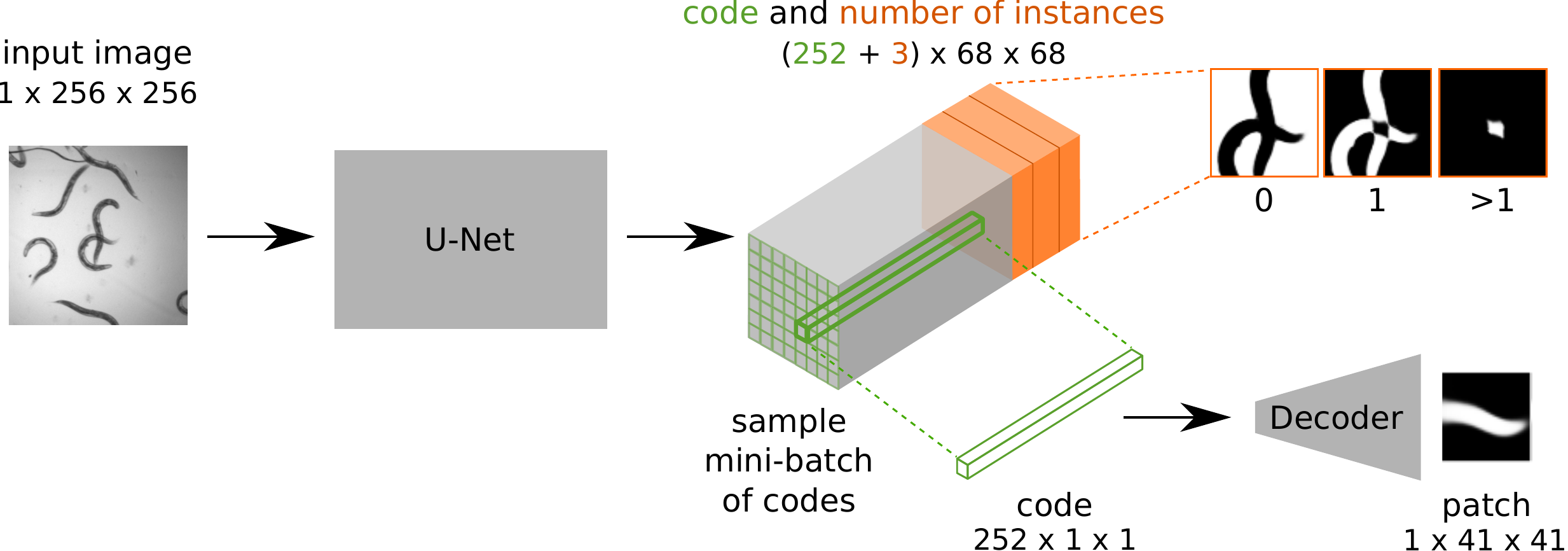}
 \caption{\label{fig:ppp_dec_architecture} \textbf{ppp+dec architecture:} A U-Net predicts shape patch encodings, which are fed into the decoding path of an auto-encoder. Additional outputs of the U-Net predict the number of instances at each pixel. U-Net and decoder are trained jointly end-to-end. Categorical cross-entropy is used for the number of instances, and binary cross-entropy for the patch predictions. Both losses are summed up without weighting. The batch of codes that is run through the decoder is sampled from pixels for which the number of instances is predicted to be 1. }
 \label{fig:arch}
\end{figure*}

Our baseline PatchPerPix architecture, termed \textbf{ppp}, is a U-Net that directly outputs $p_{\mathbf{x}}$ at each pixel $\mathbf{x}$ of the input image $I$.
To estimate the image foreground $\textnormal{fg}(I)$, we include offset $\mathbf{0}$ in $\mathcal{P}$. 
A practical issue with ppp is that the size of the predicted patches, i.e.\ the number of outputs of the U-Net, is limited by GPU memory. 
Furthermore, in most application domains, the variety of possible patch predictions, and hence the amount of information contained in each, is limited.
Therefore, in addition to our baseline model, we explore two variants that learn compressed representations of $p_{\mathbf{x}}$ and decode these via (1) the decoder part of a separately trained autoencoder (ppp+ae), and
(2) a decoder that is trained end-to-end with the backbone U-Net (ppp+dec), as described in the following.

\noindent{\textbf{ppp+ae.   }}
In a separate first step, we train a fully convolutional autoencoder on patches of ground truth binary masks to learn a patch latent space.
The backbone U-Net is then trained to regress a respective learnt latent vector (a.k.a.\ "encoding" or "code") for each pixel of the input image w.r.t.\ sum of squared differences loss.
To de-compress patch predictions for our instance assembly pipeline,
the decoder part of the pre-trained autoencoder is employed.
We add an extra output channel to the U-Net to predict a foreground mask, trained w.r.t.\ cross-entropy loss and added to the code loss without any weighting. Codes are decoded for all foreground pixels obtained by thresholding the foreground mask.

\noindent\textbf{ppp+dec. }
Here, we attach the decoder part of the autoencoder used in ppp+ae to the end of the U-Net and train the resulting joint network end-to-end from scratch w.r.t.\ cross-entropy. As before, the U-Net part of the network outputs the code. However, there is no loss employed directly on the code. To fit end-to-end training onto GPU memory, we sample codes from ground truth foreground pixels at training time, which are then fed to the decoder network. Similarly to ppp+ae, we extend the U-Net to simultaneously predict the foreground to allow for decoding only foreground pixels. This architecture is depicted in Figure~\ref{fig:arch}.

ppp+dec combines a U-Net for predicting a shape encoding with the decoder part of an auto-encoder. Interestingly, this end-to-end trainable architecture has two decoding parts, namely (1) the upward path of the U-Net, which serves for combining high-level image information captured at lower layers with low-level information from upper layers, and (2) the decoder half of an auto-encoder, which is needed to decompress local shape predictions in the end. 
Furthermore, especially when dealing with sparse data, the \mbox{U-Net} performs many dispensable computations, namely on background pixels.
Hence we investigated whether our proposed architecture could be replaced by a standard encoder-decoder architecture alone, with one encoding and one decoding path, like e.g.~\cite{badrinarayanan2017segnet}, as follows: 

\noindent\textbf{ed-ppp. }
Our architecture takes an image patch the size of our shape patches plus some surrounding receptive field as input, and generates a shape patch for the respective central pixel's instance as output. It is applied in a sliding window fashion on all pixels in the image foreground. We use $3\times3$ down- and upsampling to facilitate the singling out of the center pixel's instance from its neighboring instances. 
To determine for which pixels to run the encoder-decoder network, we train a separate U-Net to generate a foreground mask in a preceding step. 

\subsection{Overlapping Regions}
\label{subsec:overlap}
The case of multiple objects sharing pixels can be found in many biomedical applications, e.g.\ in 2d images of model organisms such as worms that crawl on top of each other, or neurons in light microscopy data that share pixels due to the partial volume effect.
As pixels located in areas of overlap belong to multiple instances, their respective shape patch is not well-defined. Hence we exclude these pixels from the entire pipeline.
During training, we achieve this by masking out these areas in the loss computation.
To detect overlap at test time, we predict the number of instances per pixel by extending the foreground classification task by an "overlap" class, which, as before, is trained jointly with the patch predictions by means of added cross-entropy loss.
This information is then used in the instance assembly:
Pixels in overlapping regions are discarded, i.e.\ their respective shape patches do not contribute to the consensus and cannot be selected.
This constitutes a limitation of our method in that
\begin{enumerate*}[label=(\roman*)]
    \item only overlapping regions with a maximum diameter of smaller than the size of the patches can be covered completely by patch shapes, and
    \item only occlusions within the range of the neighborhood used in the patch graph generation can be bridged
\end{enumerate*}.


\section{Results}
We evaluate our method on four benchmark datasets, which comprise overlapping objects, sophisticated object shapes, and, to show the generic applicability of our method, also simple object shapes. 
The first dataset, the BBBC010 C.\,elegans dataset~\cite{celegans_ljosa2012}, exhibits clusters of overlapping objects with large, coinciding bounding boxes. The second dataset, the ISBI 2012 Challenge on segmenting neuronal structures in electron microscopy~\cite{arganda2015crowdsourcing}, exhibits densely clustered objects with sophisticated shapes that span the whole image, albeit without overlaps. The third and fourth dataset exhibit densely clustered objects of simple, approximately ellipsoidal shapes, namely 2d and 3d fluorescence light microscopy datasets of cell nuclei~\cite{stardist_schmidt2018,stardist3d_weigert2019}.
Our results define the new state of the art in all cases, as detailed in Sections~\ref{subsec:BBBC010}, \ref{subsec:ISBI2012}, and~\ref{subsec:nuclei}. 
Furthermore, we study the impact of individual steps of our instance assembly pipeline as well as our proposed network architecture designs on BBBC010. 
Last, we show promising qualitative results on 3d light microscopy data of neurons, which exhibit extreme cases of sophisticated object shapes that form dense clusters with overlaps (Sec. \ref{subsec:flylight}).

\subsection{BBBC010 C.\,elegans worm disentanglement}
\label{subsec:BBBC010}
The \textbf{BBBC010} dataset from the Broad Bioimage Benchmark Collection~\cite{celegans_ljosa2012}\footnote{BBBC010v1: C.\,elegans infection live/dead image set version 1 provided by Fred Ausubel} consists of 100 brightfield microscopy images showing multiple C.\,elegans worms per image, which may overlap and cluster.
As ground truth, to capture overlaps correctly, BBBC010 provides an individual binary mask for each worm. 

In this section, we report a quantitative evaluation of our method in comparison with related work~\cite{singlingout_yurchenko2017,semiconv_novotny2018,Kulikov_2020_CVPR,chen_2022_layering,waehlby2012}. Furthermore, we report a comparison of the neural network architecture designs we explored, as well as an ablation study that assesses the impact of individual steps of our instance assembly pipeline. The impact of patch- and code-size hyperparameters is studied in \ifthenelse{\boolean{my_hasappendix_check}}{Suppl.\ Sec.~\ref{suppl:wb_ablation_code_patch_size}}{Suppl.\ Sec.~C}. 

As backbone CNN architecture we employ a 4-level U-Net~\cite{unet_ronneberger2015} starting with 40 feature maps, with two-fold down- and upsampling operations, and constant number of feature maps during upsampling. 
%
%
In the ppp architecture, we employ a patch size of $25\times25$, yielding a U-Net with $625$ outputs. 
In the ppp+ae and ppp+dec architectures, we employ a code of size $252$ as intermediate output of the U-Net, which is then fed into a decoder network to yield a patch of size $41\times41$. 
The ed-ppp architecture takes $81\times81$ patches of the raw image as input and predicts $41\times41$ shape patches. It applies \(3\times3\) max-pooling three times, and has two convolutional layers on each level. At the bottleneck, the code has an extent of $3\times3\times256$ and uses $1\times1$ convolutions. The network is symmetric and uses same padding. The output is cropped to obtain the desired patch shape.
For training, we use standard augmentation including elastic deformations in all experiments. Contrary to \cite{singlingout_yurchenko2017}, we do not augment the number of worms synthetically, but focus on crowded regions during training. 
For patch graph partitioning, we explore connected component analysis on the positive subgraph (CC) as well as the mutex watershed (MWS)~\cite{mutex_wolf2018}, where MWS is our default choice. 

 If not noted otherwise, we divide the BBBC010 dataset into training- and test set with 50 images each, as in~\cite{singlingout_yurchenko2017,semiconv_novotny2018,Kulikov_2020_CVPR}. Furthermore, we apply 2-fold cross-validation on the test set to determine the number of training steps and the patch foreground threshold $t$ (Eq.\ \ref{eq:patch_fg}), individually in all experiments. 
See \appendixcheck{my_hasappendix_check}{Suppl.\ Table~\ref{tab:train_val_test_split}}{Suppl.\ Table~6} for a precise definition of our train/val/test splits.
We report results in terms of the evaluation score used in the kaggle 2018 data science bowl, S\(\ :=\frac{TP}{TP+FP+FN}\), which takes both missing and spurious instances into account~\cite{Caicedo2019}\footnote{\url{https://www.kaggle.com/c/data-science-bowl-2018}}.
We also report a range of additional metrics that have been reported for competing approaches, including AP~\cite{coco_metric2014}, for comparability.\footnote{See \ifthenelse{\boolean{my_hasappendix_check}}{Suppl.\ Sec.~\ref{suppl:analysis_ap_coco}}{Suppl.\ Sec.~H} for a discussion of pitfalls regarding the evaluation of AP on BBBC010.}

\begin{table*}[t]
	\centering
	\caption{Quantitative results on the BBBC010 dataset. We compare to competing approaches in various metrics and setups due to a missing standard: We report COCO metrics~\cite{coco_metric2014} as in~\cite{semiconv_novotny2018,Kulikov_2020_CVPR}, recall at different IoU- and Dice thresholds as in~\cite{singlingout_yurchenko2017,waehlby2012}, and  S~\cite{Caicedo2019} on subset D of BBBC010 as in~\cite{chen_2022_layering}.
          Top: we train and evaluate on binary foreground segmentation masks as in~\cite{semiconv_novotny2018,singlingout_yurchenko2017,Kulikov_2020_CVPR}. 
          Middle: we train and evaluate on raw brightfield images as in~\cite{waehlby2012,chen_2022_layering}. 
          Bottom: Results for the architecture setups we explored (trained and evaluated on raw brightfield images).
          %
        }
	\label{tab:wb_results}
	\begin{tabu} to \linewidth{ X[2.2l] X[1.0c] X[1.0c] X[0.2c] X[1.0c] X[1.0c] X[1.0c]}
		\toprule
		\multicolumn{7}{c}{BBBC010}\\
		\midrule
                   Comparative Evaluation, Input: Binary fg masks                        & Recall$_{\textnormal{IoU}\geq0.5}$ & Recall$_{\textnormal{IoU}\geq0.8}$   &     & avAP$_{[0.5:0.05:0.95]}$    & AP$_{0.5}$      & AP$_{0.75}$    \\
 \midrule
 SON \cite{singlingout_yurchenko2017}      & $\sim$ 0.97    & $\sim$ 0.7        &     & -                                       & -              & -              \\
 Semi-conv Ops \cite{semiconv_novotny2018} & -              & -          &     & 0.569                                   & 0.885          & 0.661          \\
 Harmonic Emb. \cite{Kulikov_2020_CVPR}    & -              & -               &     & 0.724                                   & 0.900          & 0.723          \\
 PatchPerPix~(ppp+dec)                     & \textbf{0.988} & \textbf{0.964} & & \textbf{0.925}                          & \textbf{0.977} & \textbf{0.936} \\
                \midrule
                \midrule
       \end{tabu}
       \begin{tabu} to \linewidth{ X[4.5l] X[2.0c] X[0.1c]  X[2.5c]  X[2.5c] X[1.0c] X[1.0c] X[1.0c] X[1.0c] X[1.0c] X[1.0c]}
                Comparative Evaluation, & Recall$_{\textnormal{Dice}\geq0.8}$ & & avS$_{[0.5:0.05:0.95]}$ & avS$_{[0.5:0.1:0.9]}$ & S$_{0.5}$ & S$_{0.6}$ & S$_{0.7}$ & S$_{0.8}$ & S$_{0.9}$\\
                Input: Images & & & \multicolumn{7}{l}{
                	\begin{tikzpicture}
    			\def\x{0.4}
   			 \def\tick{0.1}
   			 \draw (0,0) -- (\x,0);
    			\draw (0,\tick) -- ++(0,-2*\tick) ;
		\end{tikzpicture}
		 eval.\ on train/test split from~\cite{chen_2022_layering} for comparability, \appendixcheck{my_hasappendix_check}{cf.\ Suppl.\ Tab.\ \ref{tab:train_val_test_split}}{Suppl.\ Tab.\ 6} 
                	\begin{tikzpicture}
    			\def\x{0.4}
   			 \def\tick{0.1}
   			 \draw (0,0) -- (\x,0);
    			\draw (\x,\tick) -- ++(0,-2*\tick) ;
		\end{tikzpicture}
 }\\
		\midrule
		 WormToolbox \cite{waehlby2012}            & 0.81 &  & - & -   & -   & -   & -   & -   & -   \\
                Inst.\,Seg.\,via Layering \cite{chen_2022_layering} & - &  & - & 0.754 & 0.936 & 0.919 & 0.865 & 0.761 & 0.290\\
                PatchPerPix~(ppp+dec) & \textbf{0.978} & & 0.761 & \textbf{0.816} & \textbf{0.960} & \textbf{0.955} & \textbf{0.931} & \textbf{0.805} & \textbf{0.428}\\
	 \midrule
         \midrule
	\end{tabu}
       \begin{tabu} to \linewidth{ X[4.5l] X[2.0c] X[0.1c] X[2.5c]  X[2.5c] X[1.0c] X[1.0c] X[1.0c] X[1.0c] X[1.0c] X[1.0c]}
         \multicolumn{2}{l}{Evaluation of Architecture Setups}  & & avS$_{[0.5:0.05:0.95]}$ & avS$_{[0.5:0.1:0.9]}$ & S$_{0.5}$ & S$_{0.6}$ & S$_{0.7}$ & S$_{0.8}$ & S$_{0.9}$\\
		\midrule
		ppp & & & 0.689 & 0.737 & 0.890 & 0.872 & 0.840 & 0.710 & 0.372\\
		ppp+ae & & & 0.617 & 0.660 & 0.878 & 0.831 & 0.783 & 0.610 & 0.199\\
		ppp+dec & & &\textbf{0.727} & \textbf{0.778} & \textbf{0.930} & \textbf{0.905} & \textbf{0.879} & \textbf{0.792} & \textbf{0.386}\\
		ed-ppp & & & 0.675 & 0.721 & 0.891 & 0.853 & 0.820 & 0.734 & 0.309\\
                \bottomrule
       \end{tabu}
      \end{table*}
\begin{figure*}
	\centering
	\def\svgwidth{0.81\textwidth}
	\small{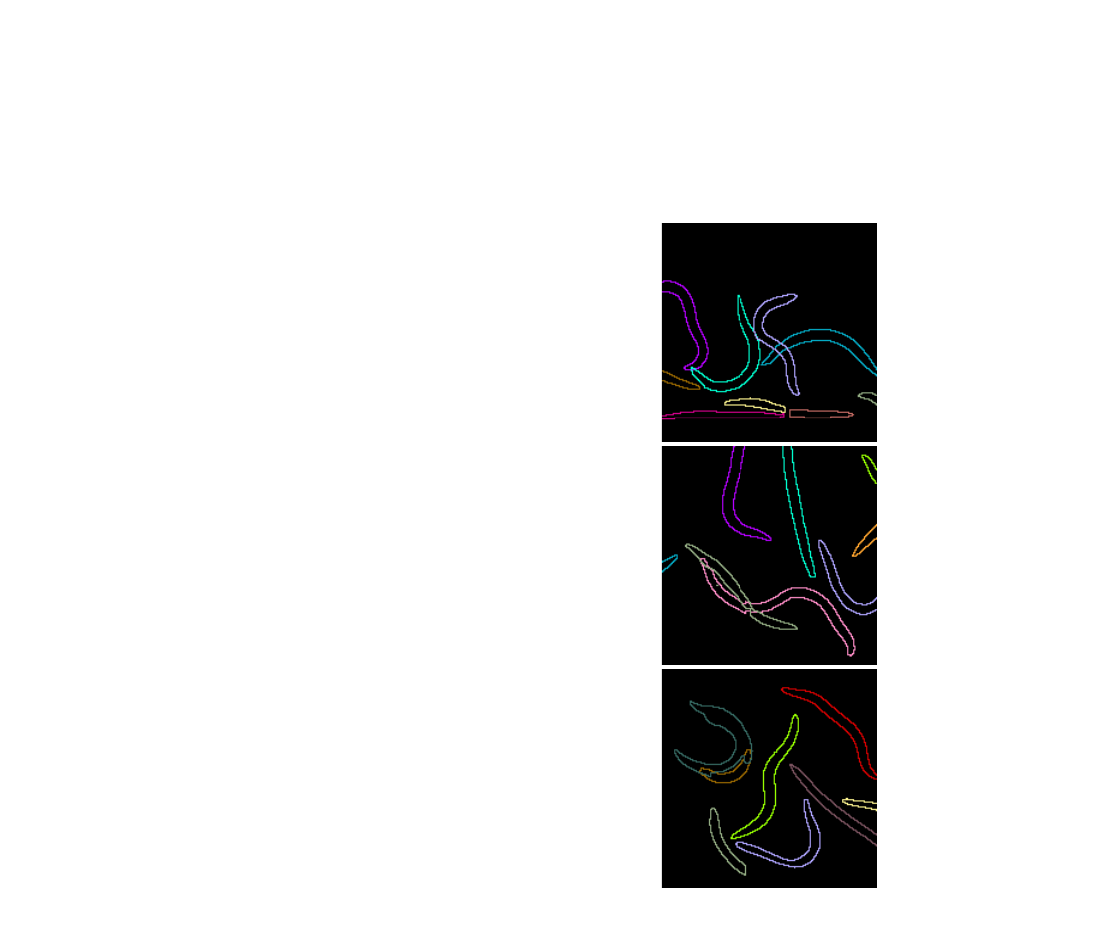}
	\caption{Qualitative results for exemplary challenging regions of the BBBC010 dataset. All architectures are able to handle crowded and overlapping regions, where ppp+dec yields fewest errors. However, rare shapes such as very bent worms are segmented with slightly higher accuracy by ppp. }
    \label{fig:wb_qualitative}
\end{figure*}
Table~\ref{tab:wb_results} compares state-of-the-art methods~\cite{semiconv_novotny2018,singlingout_yurchenko2017,Kulikov_2020_CVPR,chen_2022_layering} and PatchPerPix variants. 
Table~\ref{tab:wb_ablation} lists results of our ablation study. 
Figure~\ref{fig:wb_qualitative} shows exemplary PatchPerPix results for different CNN architectures.
\ifthenelse{\boolean{my_hasappendix_check}}{Suppl.\ Fig.~\ref{fig:comparison_son}}{Suppl.\ Fig.~1}
compares PatchPerPix with Singling Out Networks~\cite{singlingout_yurchenko2017} on an exemplary image.

PatchPerPix improves over competing methods by a considerable margin (cf.\ Table~\ref{tab:wb_results}, top and middle). Singling Out Networks (SON~\cite{singlingout_yurchenko2017}) are of limited pixel accuracy by design, hence superior performance of PatchPerPix at high IoU thresholds is no surprise. However, PatchPerPix is not just more pixel accurate, but outperforms SON across the IoU threshold range. PatchPerPix also outperforms Harmonic Embeddings~\cite{Kulikov_2020_CVPR}, a metric learning variant that amends the restricted pixel accuracy of SON, yet struggles at disentangling dense clusters of worms. 
          Finally, PatchPerPix outperforms Instance Segmentation via Layering~\cite{chen_2022_layering}, an approach for spontaneously assigning objects to different output channels to eliminate crowdedness, across thresholds in a direct comparison on their train/test split.

Interestingly, ppp+dec does not just outperform the separately trained ppp+ae, but also outperforms ed+ppp. I.e., \emph{using a full U-Net as an encoder,} followed by a standard decoder, considerably outperforms a standard encoder-decoder architecture applied in a sliding-window fashion (cf.\ Table~\ref{tab:wb_results}, bottom). 

Our ablation study (Table~\ref{tab:wb_ablation}) shows the significant impact of two core ideas of our instance assembly pipeline, namely consensus affinity computation (absent in MWS-Dense, avS -0.176) and selecting a sparse set of high-ranking patch predictions while culling low-ranking ones by means of consensus agreement scores (absent in ppp+dec w/o selection, avS - 0.131).
These scores correlate significantly with true patch quality (cf.\ \ifthenelse{\boolean{my_hasappendix_check}}{Suppl.\ Fig.~\ref{fig:correlation_score_iou_suppl}}{Suppl.\ Fig.~4}). Thinning out a pre-selection of high ranking patches has a small impact on accuracy (ppp+dec w/o thinout, avS - 0.004), yet also positively affects run-time. Patch graph partitioning via CC vs.\ MWS are on a par on the BBBC010 data.

\ifthenelse{\boolean{my_hasappendix_check}}{Suppl.\ Fig.~\ref{fig:failure_cases}}{Suppl.\ Fig.~2}
shows exemplary failure cases of ppp+dec.
Interestingly, strongly bent worms are captured with inferior pixel accuracy by our encoding-based model ppp+dec as opposed to ppp (see \ifthenelse{\boolean{my_hasappendix_check}}{Suppl.\ Fig.~\ref{fig:failure_cases}}{Suppl.\ Fig.~2}
right and Fig.~\ref{fig:wb_qualitative} top row).

\begin{table*}[htb]
    \centering
    \caption{Ablation study for PatchPerPix on the BBBC010 dataset. We ablate consensus affinity computation as a whole by running graph partitioning directly on the predictions $p$ interpreted as dense affinities (MWS-Dense). We ablate patch selection as a whole (ppp+dec w/o selection), and thinning of the patch selection (ppp+dec w/o thinout). We run ppp+dec with a standard U-Net, i.e.\ with decreasing number of feature maps in the up-sampling path (ppp+dec std U-Net). Last, we compare patch graph partitioning with CC vs.\ MWS. }
    \label{tab:wb_ablation}
    \begin{tabu} to 0.99\linewidth{ X[4.1l] X[2.3c] X[0.9c] X[0.9c]  X[0.9c] X[0.9c] X[0.9c]}
        \midrule
        S & avS$_{[0.5:0.05:0.95]}$ & S$_{0.5}$ & S$_{0.6}$ & S$_{0.7}$ & S$_{0.8}$ & S$_{0.9}$\\
        \midrule
        MWS-Dense & 0.551 & 0.687 & 0.676 & 0.661 & 0.586 & 0.326\\
        ppp+dec w/o selection & 0.596 & 0.878 & 0.853 & 0.798 & 0.544 & 0.157 \\
        ppp+dec w/o thinout & 0.723 & 0.924 & 0.898 & 0.871 & 0.788 & 0.393\\
        ppp+dec std U-Net & 0.719 & 0.916 & 0.891 & 0.873 & 0.766 & \textbf{0.406}\\
        ppp+dec, CC & 0.723 & 0.922 & 0.894 & 0.873 & 0.780 & \textbf{0.406}\\
        ppp+dec, MWS & \textbf{0.727} & \textbf{0.930} & \textbf{0.905} & \textbf{0.879} & \textbf{0.792} & 0.386\\
        \bottomrule
    \end{tabu}
\end{table*}

\subsection{ISBI 2012 neuron EM segmentation}
\label{subsec:ISBI2012}
We evaluate our method on the ISBI 2012 Challenge on segmenting neuronal structures in electron microscopy (EM) data~\cite{arganda2015crowdsourcing}. The data consists of 30 slices of 512x512 pixels with known ground truth (training data), and another 30 such slices for which ground truth is kept secret by the Challenge organizers (test data). Our network architecture as well as the training- and prediction procedure closely follows~\cite{mutex_wolf2018}, with the difference that our network has 625 instead of 17 outputs, namely patches of size 1x25x25, and we do not reduce the number of filters in the upward path of the U-Net. For partitioning the patch graph, we use the mutex watershed algorithm~\cite{mutex_wolf2018}, which has proven powerful in avoiding false mergers in case of missing neuron membrane signal in the image data. 

Our method is the leading entry on the Challenge's leaderboard\footnote{\url{http://brainiac2.mit.edu/isbi_challenge/leaders-board-new}} at present among thousands of submissions by more than 200 teams. 
Table~\ref{tab:isbi-results} lists results obtained with PatchPerPix in terms of the Challenge error metrics, robust Rand score (rRAND) and robust information theoretic measure (rINF), evaluated on the test data. For comparison, the table also lists the previous state of the art as obtained via sparse affinity predictions processed with the mutex watershed algorithm~\cite{mutex_wolf2018} (MWS). Furthermore, as an additional baseline, we interpreted our patch predictions as dense affinities which we processed with the mutex watershed algorithm as in~\cite{mutex_wolf2018} (MWS-Dense). 
PatchPerPix slightly outperforms MWS in terms of the leaderboard-defining rRAND score. This can be attributed to fewer mistakes on large neuronal bodies which have respective large impact on the rRAND score. However, the number of such large mistakes we were able to identify by eye on the test set is very small  in both approaches. 

Interestingly, MWS-Dense performs considerably worse than both PatchPerPix and MWS. The difference between MWS-Dense and PatchPerPix can be attributed to individual erroneous predictions causing errors in MWS-Dense, which are amended in PatchPerPix by our proposed consensus voting and patch selection scheme. As for the difference between MWS-Dense and MWS, we hypothesize that this is due to MWS smartly distinguishing between purely attractive short-range- and purely repulsive long-range affinities. Instead, MWS-Dense treats all affinities as both attractive and repulsive. 
\begin{table}[t]
  \centering
  \caption{Quantitative results for the ISBI 2012 Challenge on segmenting neuronal structures in electron microscopy data~\cite{arganda2015crowdsourcing}. PatchPerPix defines the current state-of-the-art in terms of the leaderboard-defining rRAND score. 
  }
	\begin{tabu} to 0.9\linewidth{ X[2.0l] X[1.0c] X[1.0c] }
		\toprule
		ISBI2012 & rRAND & rINF \\
		\midrule
		PatchPerPix & \textbf{0.988290} & 0.991544 \\
		MWS~\cite{mutex_wolf2018} & 0.987922 & \textbf{0.991833}  \\
		MWS-Dense & 0.979112 & 0.989625 \\
		\midrule
	\end{tabu}
    \label{tab:isbi-results}
\end{table}

\subsection{Nuclei segmentation in 2d and 3d}
\label{subsec:nuclei}
We evaluate our method on 2d and 3d fluorescence microscopy images of cell nuclei. 
The 2d dataset is a subset of the kaggle 2018 data science bowl\footnote{BBBC038v1: available from the Broad Bioimage Benchmark Collection\cite{celegans_ljosa2012}} as defined in~\cite{stardist_schmidt2018}. It consists of 380 training, 67 validation and 50 test images. We refer to this dataset as \textbf{dsb2018}.
The 3d dataset consists of 28 confocal microscopy images collected and annotated by~\cite{Long2009}. Image size is approximately $140 \times 140 \times 1100$ pixels. Each image shows hundreds of nuclei, with multiple dense clusters. An example is shown in \ifthenelse{\boolean{my_hasappendix_check}}{Suppl.\ Fig.~\ref{fig:nuclei3d_suppl}}{Suppl.\ Fig.~3}. We partition the data as in~\cite{stardist3d_weigert2019,hirsch20_an_auxil_task_for_learn_nucle_segme_in_3d_micro_image}, with 18 images for training, 3 for validation, and 7 for testing. We refer to this dataset as \textbf{nuclei3d}.

For dsb2018, our CNN architecture is a 4-level U-Net, with 40 initial feature maps, that predicts foreground/background labels as well as codes of size 256, decoded into patches of size $25 \times 25$.
We determine the number of training steps as well as the patch threshold on the validation set.
For nuclei3d, we employ a 3-level 3d U-Net with 20 initial feature maps, tripled after each downsampling step.
We predict patches of size \(9 \times 9 \times 9\). We filter out instances smaller than a threshold. We determine the number of training steps, the patch threshold, and the instance size threshold on the validation set.

Table~\ref{tab:nuclei_results} lists our results in comparison to the previous state of the art on this data~\cite{stardist_schmidt2018,stardist3d_weigert2019,hirsch20_an_auxil_task_for_learn_nucle_segme_in_3d_micro_image}. We furthermore compare to MALA~\cite{connectome_funke2018}, an affinity-based instance segmentation method trained with a structured loss, which is an established baseline for a different kind of 3d data, namely 3d electron microscopy of neuronal structures, but does not explicitly capture instance shape. For MALA, we employ the same backbone U-Net as for PatchPerPix for a fair comparison.

Superior avS of PatchPerPix compared to~\cite{stardist_schmidt2018,stardist3d_weigert2019} can be attributed to superior performance at high IoU thresholds, where StarDist's pixel accuracy is limited due to its coarse polyhedral shape representation, especially in 3d. We list IoU thresholds down to 0.1 as in~\cite{stardist3d_weigert2019}, indicating that PatchPerPix is on a par with StarDist in terms of topological segmentation errors like false splits and false mergers of nuclei.
Compared to a recently proposed 3-label U-Net trained with an auxiliary task~\cite{hirsch20_an_auxil_task_for_learn_nucle_segme_in_3d_micro_image}, again, the high pixel accuracy of PatchPerPix leads to slightly higher avS, while~\cite{hirsch20_an_auxil_task_for_learn_nucle_segme_in_3d_micro_image} is slightly superior at low IoU thresholds.

On dsb2018, we observed a similar improvement of ppp+dec over ppp as on BBBC010. However, this does not hold for nuclei3d, where ppp+dec did not improve over ppp. We hypothesize that encodings are less able to capture the ellipsoidal shape of nuclei at the very small 3d patch size of 9x9x9 we're bound to to achieve manageable computational performance of instance assembly in 3d (see \ifthenelse{\boolean{my_hasappendix_check}}{Suppl.\ Table~\ref{tab:inf_time_suppl}}{Suppl.\ Table~4} for run-times). This performance bottleneck constitutes a current limitation of our method on 3d data, and is subject to future work.

\begin{table*}[t]
  \centering
  \caption{Quantitative results for the nuclei datasets dsb2018 and nuclei3d. We report the 2018 Data Science Bowl score S for multiple IoU thresholds. Note that the results are not directly comparable to their leaderboard as the test set is different.
  }
    \begin{tabu} to 1.0\linewidth{ X[3.3l] X[1.5l] X[1.0l] X[1.0l] X[1.0l]  X[1.0l] X[1.0l] X[1.0l] X[1.0l] X[1.0l] X[1.0l]}
		\toprule
        S& avS & S\(_{0.1}\) & S\(_{0.2}\) & S\(_{0.3}\) & S\(_{0.4}\) & S\(_{0.5}\) & S\(_{0.6}\) & S\(_{0.7}\) & S\(_{0.8}\) & S\(_{0.9}\)\\
        & {\scriptsize[0.5:0.1:0.9]} & \\
		\toprule
		\multicolumn{11}{c}{dsb2018}\\
		\midrule
        Mask R-CNN\cite{stardist_schmidt2018} & 0.594          & -              & -              & -              & -              & 0.832          & 0.773          & 0.684          & 0.489          & 0.189          \\
        StarDist\cite{stardist_schmidt2018}   & 0.584          & -              & -              & -              & -              & 0.864          & 0.804          & 0.685          & 0.450          & 0.119          \\
        PatchPerPix                      & \textbf{0.693} & \textbf{0.919} & \textbf{0.919} & \textbf{0.915} & \textbf{0.898} & \textbf{0.868} &
\textbf{0.827}                                & \textbf{0.755} & \textbf{0.635} & \textbf{0.379}                                                                                                                        \\
		\midrule
		\midrule
		\multicolumn{11}{c}{nuclei3d}\\
		\midrule
                MALA \cite{connectome_funke2018}    & 0.381          & 0.895          & 0.887          & 0.859          & 0.803          & 0.699          & 0.605          & 0.424          & 0.166          & 0.012          \\
                StarDist 3D\cite{stardist3d_weigert2019} & 0.406          & 0.936 & 0.926 & 0.905          & \textbf{0.855}          & 0.765 & 0.647          & 0.460          & 0.154          & 0.004          \\
                3-label+cpv\cite{hirsch20_an_auxil_task_for_learn_nucle_segme_in_3d_micro_image}                              &  0.425       & \textbf{0.937}          & \textbf{0.930}          & \textbf{0.907}          & 0.848          & 0.750          & 0.641          & 0.473          & 0.224          & \textbf{0.035}          \\
                PatchPerPix                              & \textbf{0.436}          & 0.926          & 0.918          & 0.900          & 0.853          & \textbf{0.766}          & \textbf{0.668}          & \textbf{0.493}          & \textbf{0.228}          & 0.027          \\
		\bottomrule
	\end{tabu}
    \label{tab:nuclei_results}
\end{table*}

\subsection{Neuron separation in 3d light microscopy data}
\label{subsec:flylight}

\begin{figure*}[t]
    \centering
    \begin{subfigure}{0.24\textwidth}
		\includegraphics[width=\linewidth]{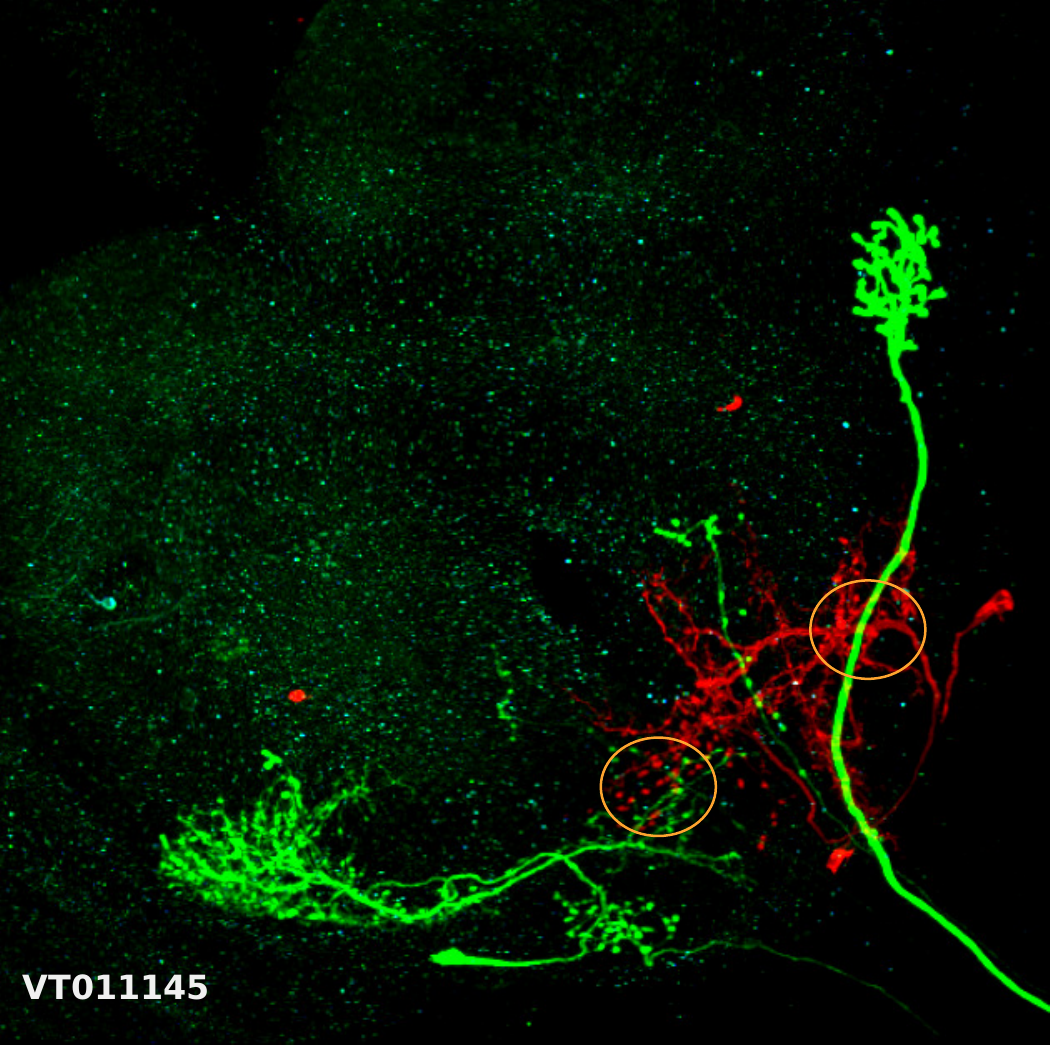}\\
		\includegraphics[width=\linewidth]{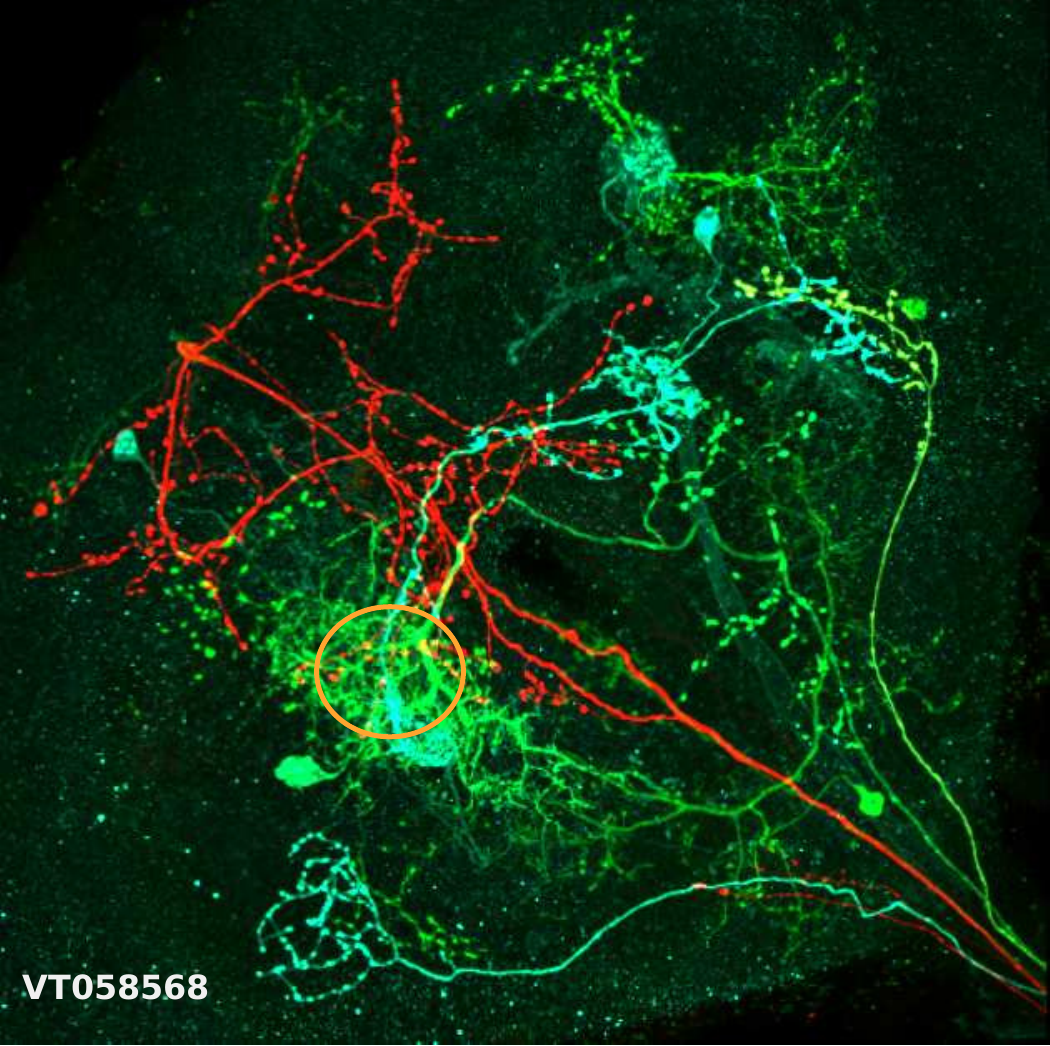}
		\caption{Raw}
	\end{subfigure}
    \begin{subfigure}{0.24\textwidth}
		\includegraphics[width=\linewidth]{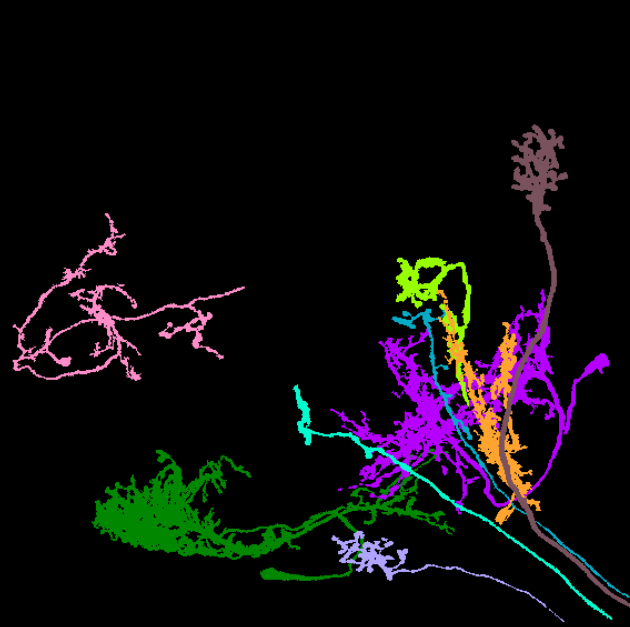}\\
		\includegraphics[width=\linewidth]{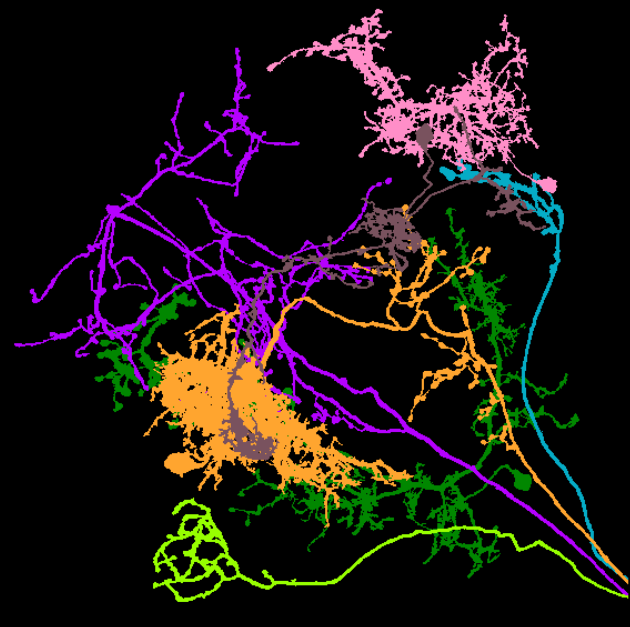}
		\caption{Ground Truth}
	\end{subfigure}
    \begin{subfigure}{0.24\textwidth}
		\includegraphics[width=\linewidth]{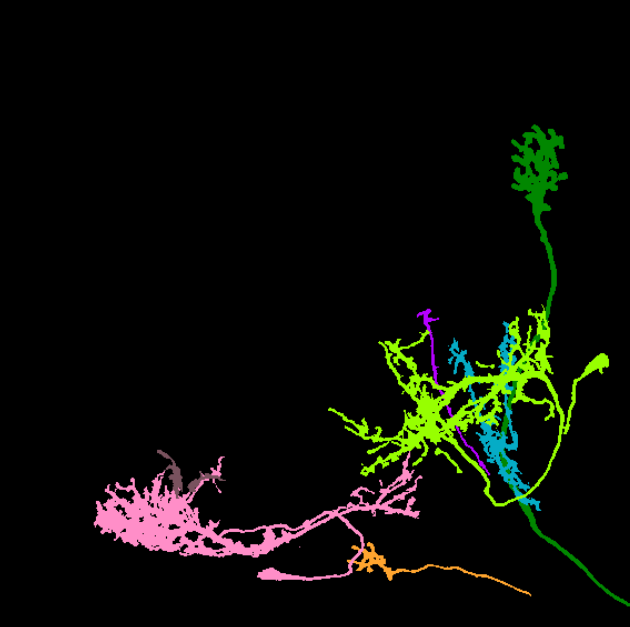}\\
		\includegraphics[width=\linewidth]{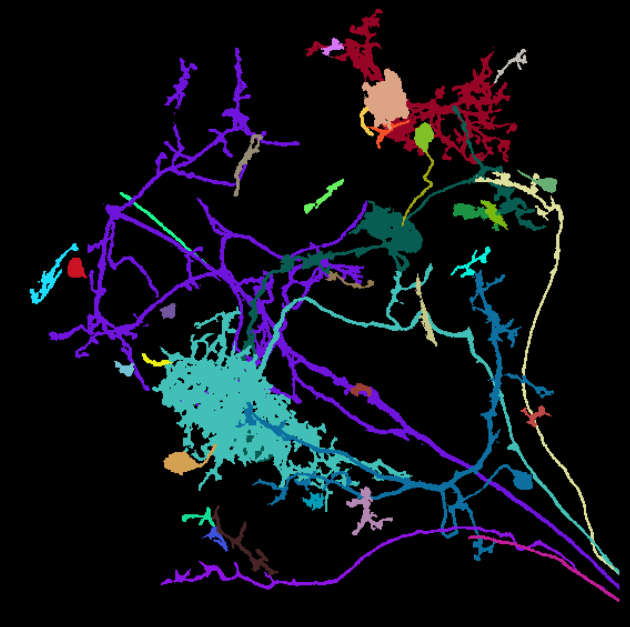}
		\caption{PatchPerPix}
	\end{subfigure}
	\caption{Qualitative results on 3d neuron light microscopy examples. (a) Maximum intensity projection of raw images. Orange circles indicate overlapping areas in 3d. Ground truth data (b) were generated by manual segmentation using VVD Viewer. PatchPerPix (c) shows promising results on this challenging dataset.}
    \label{fig:neuron_qualitative}
\end{figure*}

We aim to identify and segment neurons of the fruit fly brain (GAL4 lines~\cite{Jenett2012}) in an unpublished dataset of 3d multicolor confocal microscopy images. The imaging is done by stochastic labeling able to express different densities of neurons~\cite{mcfo_Nern2015} (cf.\ Fig.\ \ref{fig:neuron_qualitative}a). 
This instance segmentation task is very challenging as the number of neurons can be high and image quality is bounded by the necessity to perform large-scale imaging.
Moreover, the neurons are very thin, tree-like structures which are intertwined and may overlap due to partial volume effects.

As this dataset is still in the process of being curated and extended, and no competing approach has yet been reported, we do not perform a quantitative evaluation of PatchPerPix, but show the quality of exemplary results on a test set of two images in Figure~\ref{fig:neuron_qualitative}. 
We use a 3-level 3d U-Net with $2\times$ down- and upsampling and 12 initial feature maps, tripled at each downsampling. The predicted patches are of size $7\times7\times7$ pixels.
Our results serve as proof-of-concept that our method is applicable and yields reasonable results for thin, complex tree-like structures in large 3d image volumes. 


\section{Conclusion}
In this work we present a novel generic method for instance segmentation that comprises a CNN to predict dense local shape descriptors and a one-pass instance assembly pipeline. 
The method is able to handle objects of sophisticated shapes that appear in dense clusters with overlaps, including crossovers. 
It is the first to assemble all instances from learnt shape patches, simultaneously in one pass.
We successfully applied our method to a range of domains, showing that it
(1)~outperforms the state of the art on the heavily contested ISBI 2012 challenge on neuron segmentation in electron microscopy, 
(2)~outperforms the state of the art on the challenging BBBC010 C.\,elegans worm data by a large margin,
(3)~outperforms the state of the art on 2d and 3d fluorescence microscopy data of densely clustered cell nuclei (on par in terms of cell detection performance, better in terms of pixel accuracy), 
showing that our method performs well also for simple (blob-like) instance shapes, and
(4)~can be applied to extreme cases of instance shapes, like neurons in 3d fluorescence microscopy. 
Future work will tackle a performance bottleneck that becomes relevant on 3d data, where we're currently restricted to patch sizes that are most probably sub-optimally small.


\vspace{0.5em}
\noindent\textbf{Acknowledgments. }
We wish to thank Constantin Pape for his invaluable help in reproducing the training- and prediction setup from~\cite{mutex_wolf2018}, Carolina Waehlby for help with the BBBC010 data, Stephan Saalfeld and Carsten Rother for inspiring discussions, the FlyLight Project Team\footnote{\url{https://www.janelia.org/project-team/flylight}} at Janelia Research Campus for providing unpublished data, and Claire Managan and Ramya Kappagantula (Janelia Project Technical Resources) for their conscientious manual neuron segmentations.
P.H., L.M. and D.K. were funded by the Berlin Institute of Health and the Max Delbrueck Center for Molecular Medicine. P.H. was funded by HFSP grant RGP0021/2018-102. P.H., L.M. and D.K. were supported by the HHMI Janelia Visiting Scientist Program. VVD Viewer\footnote{\url{https://github.com/takashi310/VVD_Viewer}} is an open-source software funded by NIH grant R01-GM098151-01.

{
  \bibliographystyle{splncs04}
  \bibliography{main}
}

\ifthenelse{\boolean{my_hasappendix_check}}{
\clearpage
\appendix
\onecolumn
\section*{PatchPerPix for Instance Segmentation: Supplement}
\setcounter{figure}{0}
\setcounter{table}{0}
\renewcommand{\thesubsection}{\thesection\Alph{subsection}}
\subsection{Supplemental Figure: Overview of Instance Assembly}
\renewcommand{\textfraction}{0.15}
\begin{figure*}[htbp]
	\centering
	\includegraphics[width=\textwidth]{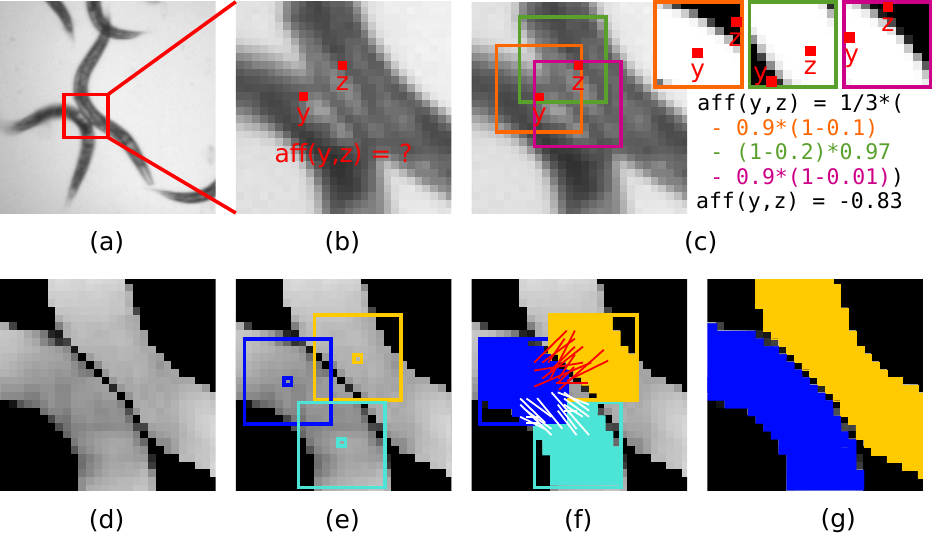}
	\caption{\label{fig:pipeline_detail_suppl} Overview of our instance assembly pipeline: (a) Raw image detail showing two closely adjacent objects,
		(b) Zoom into ambiguous area. We use two pixels, y and z, delineated in red, to visualize how the consensus affinities aff(y,z) are computed.
		(c) Three exemplary patches, marked by different-colored squares, that each cover y and z. The actual value each contributes to aff(y,z) is stated next to the image (cf.\ Eq.~3).
		(d) Patch scores for all pixels, visualized as gray value image. Predictions closer to an ambiguous region between two objects, which agree less with the consensus affinities, receive lower patch scores (cf.\ Eq.~5).
		(e) Selection of high-scoring patches that contribute to cover the image foreground (for the sake of clarity only a subset is shown). They are depicted as an overlay, where the spatial extension of each selected patch is delineated by a colored box and the center pixel by a small square.
		(f) The crayoned areas show the foreground area covered by each patch. We compute patch affinities between overlapping patches (cf.\ Eq.~6), visualized by the lines (for the sake of clarity only a subset is shown). For the two patches connected by white lines the union of foregrounds agrees well with the consensus, they belong to the same object. For the two patches connected by red lines, the union of foregrounds does not agree well with the consensus, they belong to different objects.
		(g) The final instance segmentation.
	}
\end{figure*}

\clearpage
\subsection{Supplemental Figures for BBBC010 C.\,elegans worm disentanglement}
 \begin{figure*}[htbp]
   \centering
 \begin{tabular}{cccc}
   \includegraphics[width=0.20\linewidth]{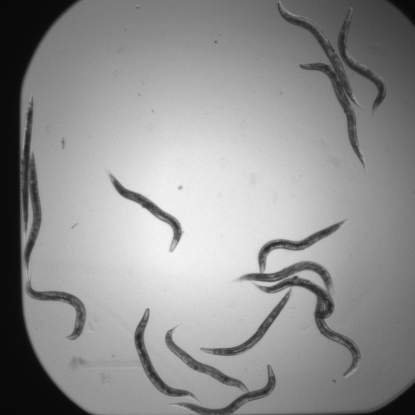} &
   \includegraphics[width=0.20\linewidth]{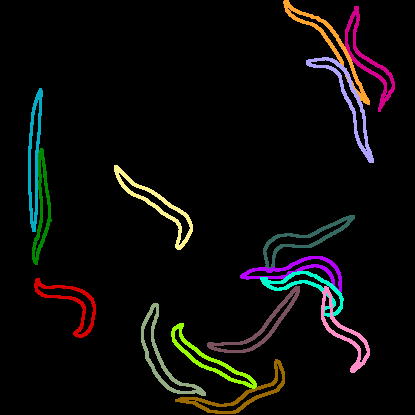}  &
   \includegraphics[width=0.20\linewidth]{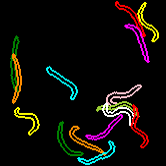}  &
   \includegraphics[width=0.20\linewidth]{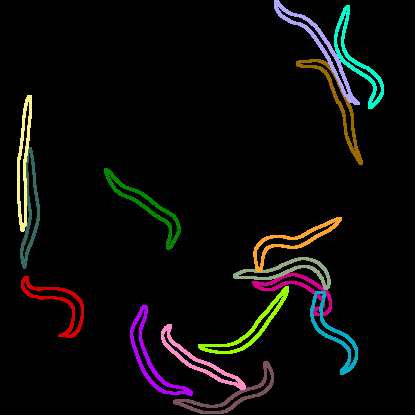}  \\
   (a) raw & (b) ground truth & (c) SON & (d) PatchPerPix
   \end{tabular}
   \caption{\label{fig:comparison_son} Qualitative comparison of PatchPerPix and Singling Out Networks~\cite{singlingout_yurchenko2017} (SON) on a BBBC010 image. PatchPerPix (ppp+dec) is significantly more pixel-accurate  as it does not rely on a dictionary of known shapes. In particular, it accurately separates a cluster of objects on the lower right. (SON image from~\cite{singlingout_yurchenko2017})}
 \end{figure*}
 \begin{figure*}[htbp]
   \centering
   \includegraphics[width=0.14\linewidth]{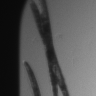}
   \includegraphics[width=0.14\linewidth]{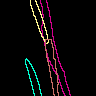}
   \includegraphics[width=0.14\linewidth]{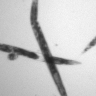}
   \includegraphics[width=0.14\linewidth]{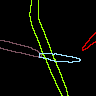} \\
   \includegraphics[width=0.14\linewidth]{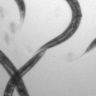}
   \includegraphics[width=0.14\linewidth]{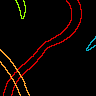}
   \includegraphics[width=0.14\linewidth]{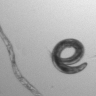}
   \includegraphics[width=0.14\linewidth]{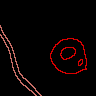}
   \caption{\label{fig:failure_cases}Exemplary failure cases of PatchPerPix (ppp+dec) on BBBC010. (top) false split due to large overlap; false split due to missing signal. (bottom) false merge due to sequential layout of worms; inaccuracy due to strongly bent worm. }
 \end{figure*}

\subsection{Supplemental Hyperparameter Studies on BBBC010}
\label{suppl:wb_ablation_code_patch_size}
\begin{table*}[htbp]
  \centering
  \caption{Impact of code size used in ppp+dec, assessed on BBBC010. Patch size fixed at 41x41, number of parameters kept constant. ppp+dec achieves comparable results across a range of compression rates we assessed.}
	\begin{tabu} to 0.99\linewidth{ X[1.5l] X[1.5c] X[2.5c] X[1c] X[1c]  X[1c] X[1c] X[1c]}
		\midrule
		S & code size & avS$_{[0.5:0.05:0.95]}$ & S$_{0.5}$ & S$_{0.6}$ & S$_{0.7}$ & S$_{0.8}$ & S$_{0.9}$\\
		\midrule
		ppp+dec & 324 & \textbf{0.737} & 0.931 & \textbf{0.919} & \textbf{0.897} & 0.784 & 0.418 \\
		ppp+dec & 252 &0.727 & 0.930 & 0.905 & 0.879 & 0.792 & 0.386\\
		ppp+dec & 216 & 0.733 & 0.924 & 0.899 & 0.878 & \textbf{0.794} & \textbf{0.425} \\
		ppp+dec & 180 & 0.730 & 0.924 & 0.906 & 0.888 & 0.786 & 0.413 \\
		ppp+dec & 144 & 0.719 & 0.912 & 0.886 & 0.868 & 0.774 & 0.410 \\
		ppp+dec & 108 & 0.734 & \textbf{0.932} & 0.914 & 0.893 & 0.776 & 0.409 \\
		ppp+dec & 72 & 0.728 & 0.923 & 0.902 & 0.881 & 0.780 & 0.412 \\
		\bottomrule
	\end{tabu}
	\label{tab:wb_ablation_code_size}
\end{table*}
\ifthenelse{\boolean{my_hasappendix_check}}{\clearpage}{}
\begin{table*}[htbp]
  \centering
  \caption{Impact of patch size used in ppp+dec, assessed on BBBC010. All ppp+dec models have similar code size of $\sim$250, as well as a comparable number of parameters. For ppp+dec, S at lower thresholds, as well as avS, tends to increase with larger patch size. This is expected, as larger patches may bridge larger overlaps of instances. However, S the highest threshold tends to decrease with larger patch size. This cannot be straightforwardly attributed to higher compression rate, as S is robust to code size variation at fixed patch size (see Suppl.\ Table~\ref{tab:wb_ablation_code_size}). We hypothesize that it is due to shape variance increasing with increasing distance from the center pixel of a patch, causing larger patches to yield lower pixel accuracy at the fringes. Furthermore, ppp+dec considerably outperforms ppp at the same patch size (25x25) and same number of parameters. Improvement of ppp+dec over ppp is largest for S at high thresholds. We hypothesize that this may be due to differences in training procedures, where in ppp+dec, only forground patches contribute to the loss, whereas in ppp, all patches contribute, thereby significantly shifting balance. Said hypotheses have to be verified by further experiments. }
	\begin{tabu} to 0.99\linewidth{ X[1.5l] X[1.5c] X[2.5c] X[1c] X[1c]  X[1c] X[1c] X[1c]}
		\midrule
		S & patch size & avS$_{[0.5:0.05:0.95]}$ & S$_{0.5}$ & S$_{0.6}$ & S$_{0.7}$ & S$_{0.8}$ & S$_{0.9}$\\
		\midrule
		ppp & 25x25 & 0.689 & 0.890 & 0.872 & 0.840 & 0.710 & 0.372 \\
		ppp+dec & 25x25 & 0.720 & 0.895 & 0.877 & 0.857 & 0.763 &  \textbf{0.450}\\
		ppp+dec & 31x31 & 0.725 & 0.911 & 0.894 & 0.866 & 0.779 & 0.417 \\
		ppp+dec & 41x41 & 0.727 & \textbf{0.930} & 0.905 & 0.879 & 0.792 & 0.386\\
		ppp+dec & 49x49 & \textbf{0.736} & 0.928 & \textbf{0.914} & \textbf{0.898} & \textbf{0.802} & 0.409 \\
		\bottomrule
	\end{tabu}
	\label{tab:wb_ablation_patch_shape}
\end{table*}

\ifthenelse{\boolean{my_hasappendix_check}}{}{\FloatBarrier}
\subsection{Supplemental Hyperparameter Study on DSB2018}
\begin{table*}[htbp]
  \centering
  \caption{Impact of patch size and code size assessed on the 2d nuclei dataset dsb2018. In line with the respective study on BBBC010, results suggest that with a smaller patch size the network is able to better learn the exact instance shape (better at high IoU thresholds) yet with a larger patch size the detection performance improves (better at smaller IoU thresholds).}
  \begin{tabu} to 1.0\linewidth{ X[1.3l] X[0.9c] X[0.9c] X[1.3l] X[1.0l] X[1.0l] X[1.0l]  X[1.0l] X[1.0l] X[1.0l] X[1.0l] X[1.0l] X[1.0l]}
    \toprule
    S& patch size & code size & avS {\scriptsize[0.5:0.1:0.9]} & S\(_{0.1}\) & S\(_{0.2}\) & S\(_{0.3}\) & S\(_{0.4}\) & S\(_{0.5}\) & S\(_{0.6}\) & S\(_{0.7}\) & S\(_{0.8}\) & S\(_{0.9}\)\\
    \midrule
    ppp     & 25     & /     & 0.670 & 0.905 & 0.905 & 0.901 & 0.882 & 0.846 & 0.797   & 0.737 & 0.603 & 0.365 \\ 
    ppp+dec & 25     & 252     & \textbf{0.693} & 0.919 & 0.919 & 0.915 & 0.898 & 0.868 & \textbf{0.827}   & 0.755 & \textbf{0.635} & \textbf{0.379} \\ 
    ppp+dec & 25     & 512     & 0.691 & 0.929 & 0.927 & 0.925 & 0.913 & \textbf{0.874} & 0.825   & \textbf{0.763} & 0.626 & 0.368 \\ 
    ppp+dec & 41     & 252     & 0.682 & 0.924 & 0.921 & 0.919 & 0.898 & 0.871 & 0.824   & 0.744 & 0.613 & 0.359 \\ 
    ppp+dec & 41     & 576     & 0.685 & \textbf{0.934} & \textbf{0.934} & \textbf{0.931} & \textbf{0.916} & 0.871 & \textbf{0.827}   & 0.750 & 0.614 & 0.361 \\ 
    \bottomrule
  \end{tabu}
  \label{tab:dsb2018_ablation_code_size_suppl}
\end{table*}

\subsection{Supplemental Study of Instance Assembly Run-times}
\begin{table*}[htbp]
	\centering
	\caption{Average run-times for PatchPerPix instance assembly on the different datasets. We achieve convenient run-times with fair patch sizes on 2d data with sparse foreground (BBBC010 and dsb2018). Long run-time for ISBI2012 at fair patch size is due to the dense foreground of the data. Long run-time at small patch size for nuclei3d is due to the data being 3d, albeit with sparse foreground. }
	\begin{tabu} to 1.0\linewidth{ X[1.5l] X[1.0c] X[1.0c]  X[1.0c] X[1.0c] X[1.0c]}
		\toprule
		dataset & BBBC010 & BBBC010 & dsb2018 & ISBI2012 & nuclei3d\\
		\midrule
		patch size & 25x25 & 41x41 & 25x25 & 25x25 & 9x9x9\\
		seconds per image & 4 & 13 & 5 & 400 & 1300\\
		\bottomrule
	\end{tabu}
	\label{tab:inf_time_suppl}
\end{table*}
\FloatBarrier

\clearpage
\subsection{Supplemental Figure for Nuclei3d}
\begin{figure*}[htb]
  \centering
  \includegraphics[width=\textwidth]{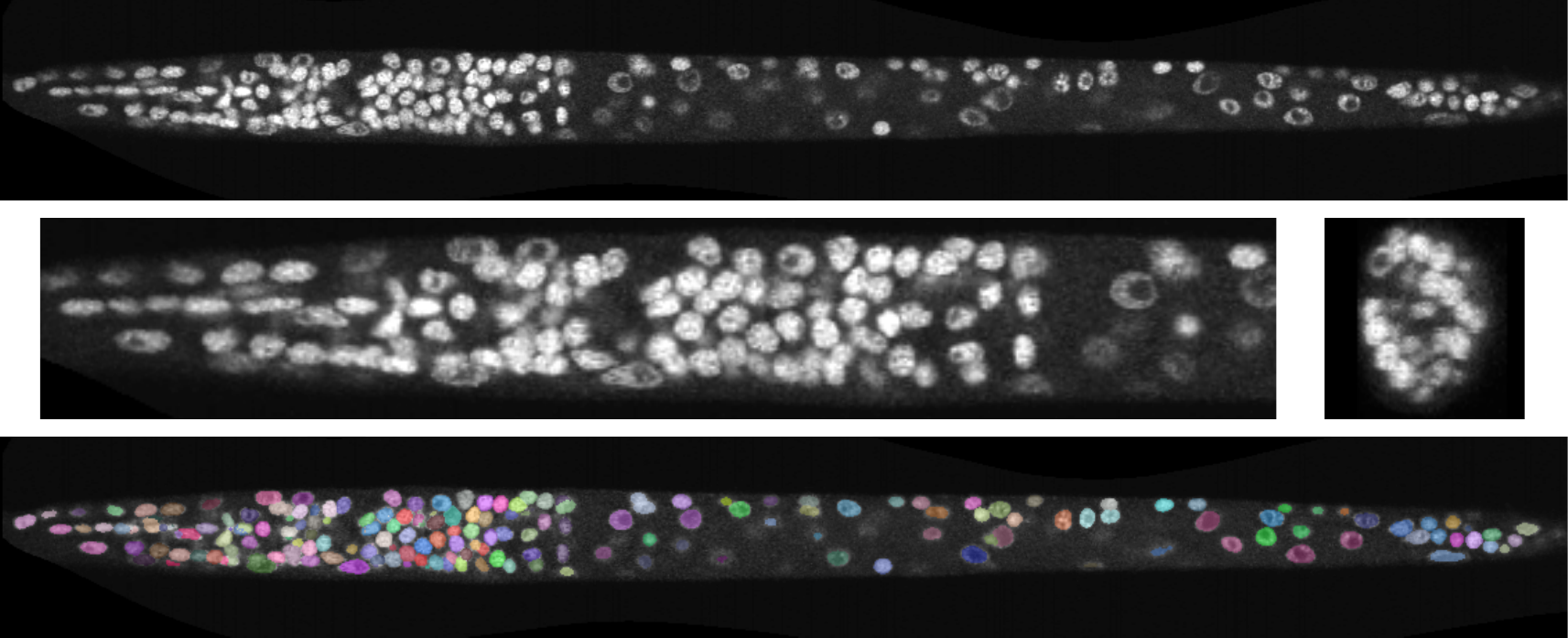}
  \caption{\label{fig:nuclei3d_suppl} Top: Exemplary xy-slice of a volume in the nuclei3d data set. Densely packed nuclei
  in the nervous system of the C.\,elegans L1 larva (towards the left)
  are particularly hard to separate.
  Center left: Close-up on said nervous system. Center right: Exemplary yz-slice of nervous system.
    Bottom: Respective PatchPerPix segmentation result.}
\end{figure*}
\FloatBarrier

\subsection{Supplemental Analysis of Patch Scores}
\begin{figure*}[htbp]
    \centering
    \begin{subfigure}{0.42\textwidth}
      \includegraphics[width=\linewidth]{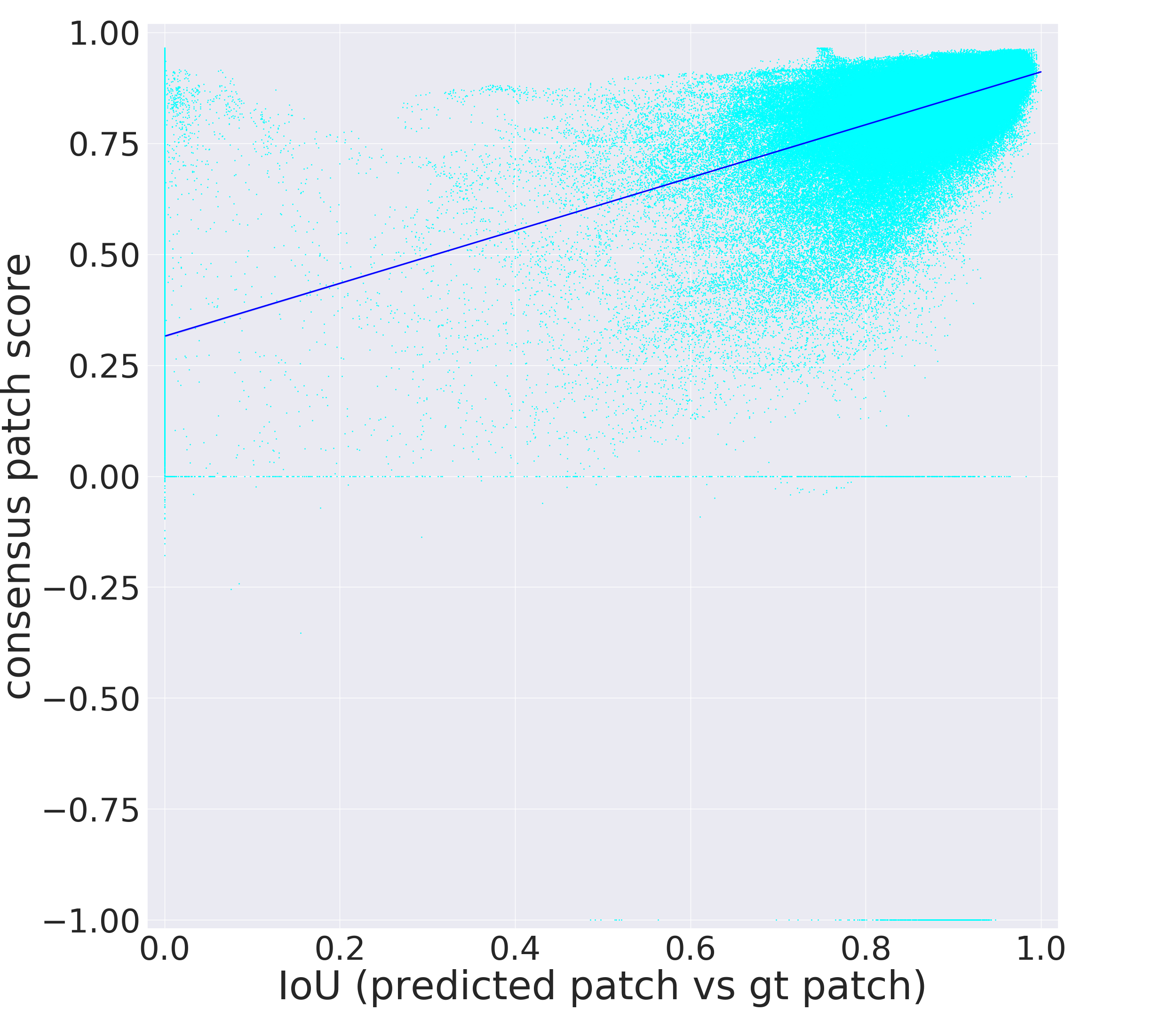}
      \caption{BBBC010}
    \end{subfigure}
    \begin{subfigure}{0.42\textwidth}
      \includegraphics[width=\linewidth]{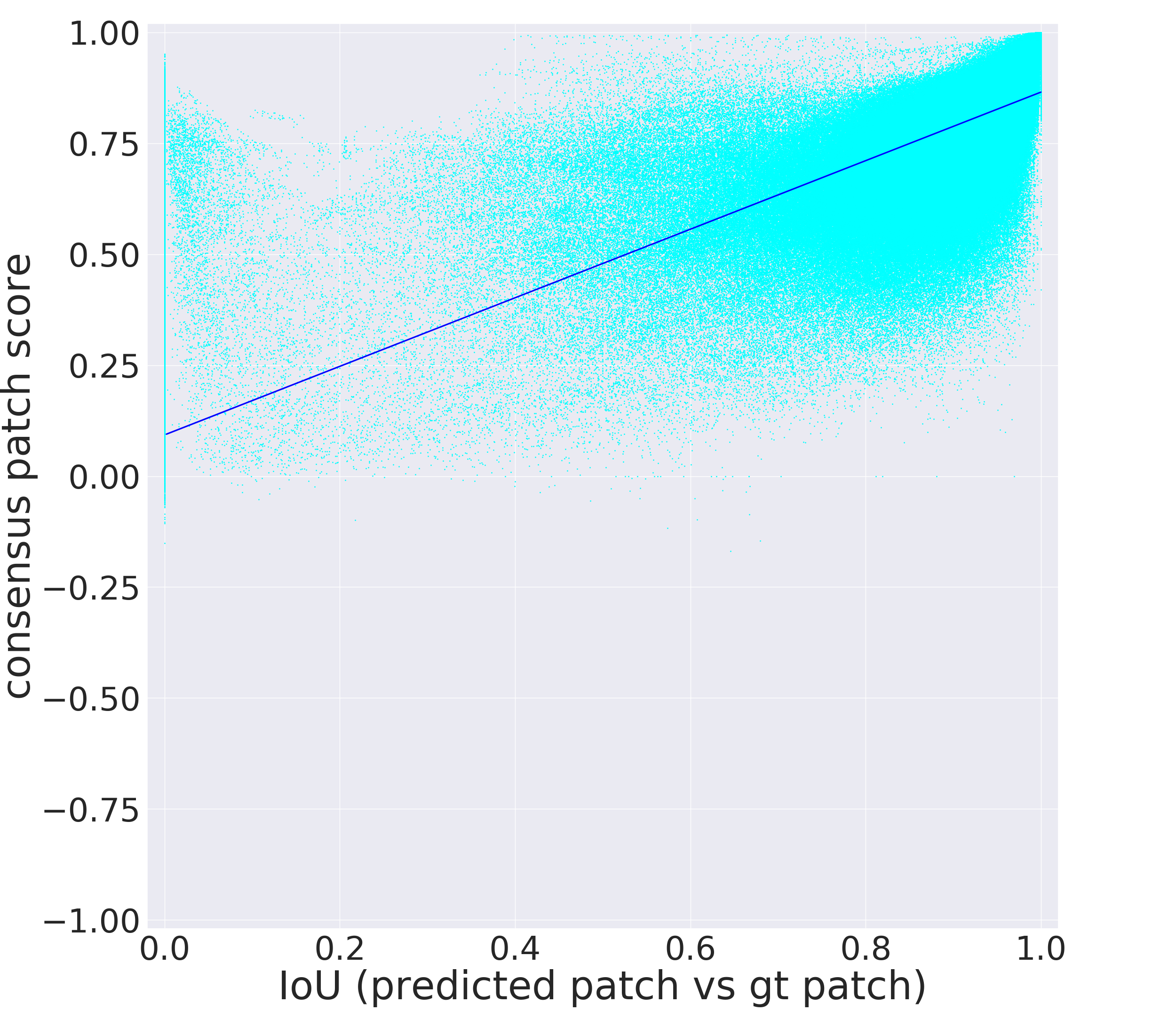}
      \caption{dsb2018}
    \end{subfigure}
    \caption{We find significant correlation between the computed patch scores and the IoU of predicted patches vs.\ ground truth patches. Each plotted dot stems from a patch prediction at a pixel that is foreground either in the prediction or in the ground truth. Correlation values: 0.52 (Spearman's rho) and 0.37 (Kendall's tau) for BBBC010, and 0.8 (Spearman's rho) and 0.62 (Kendall's tau) for dsb2018.}
    \label{fig:correlation_score_iou_suppl}
  \end{figure*}

\clearpage
\subsection{Discussion of Evaluation Metrics for BBBC010}
\label{suppl:analysis_ap_coco}

Average Precision (AP), more specifically interpolated AP~(see e.g.\ \cite{manning2008introduction}), is widely used as evaluation metric for object detection and instance segmentation, e.g. in the COCO~\cite{coco_metric2014} and CityScapes~\cite{cordts2016cityscapes} benchmarks. Interpolated AP is defined as the area under the interpolated precision-recall curve. To this end, precision-recall values are obtained at all possible confidence score thresholds. Interpolation then ensures that the precision-recall curve is well-defined and has non-positive slope everywhere.
As opposed to proposal-based methods like e.g.\ Mask RCNN~\cite{maskrcnn_he2017}, proposal-free methods, like {PatchPerPix}, do not predict a confidence score per predicted instance. In this case, by definition, interpolated AP simplifies from the area under the interpolated precision-recall curve to precision \(\boldsymbol{\cdot}\) recall.

Unfortunately, multiple widely used implementations of interpolated AP contain bugs that, while largely negligible in the common use case of existing confidence scores, have a considerable effect in case of non-existing (or all-equal) confidence scores~\cite{maier_hein_2022_metrics_reloaded}. First, the CityScapes implementation\footnote{\url{https://github.com/mcordts/cityscapesScripts}} of (exact) interpolated AP erroneously interpolates precision~$=1$ at recall~$=0$, and thus may substantially overestimate AP in the case of no confidence scores. Second, the COCO implementation\footnote{\url{https://github.com/cocodataset/cocoapi}} of 101-point interpolated AP, while interpolating correctly, processes instances with identical scores sequentially instead of computing a single precision-recall point (to be then extrapolated), and thus may also overestimate AP. (On the other hand, correct 101-point interpolated AP is smaller or equal to exact interpolated AP, yet the approximation is guaranteed to lie within a 1\% bound.) Supp.\ Table~\ref{tab:wb_suppl_results} assesses the impact of different (erroneous and correct) implementations of AP empirically on BBBC010 in the "Input: Images" setup also evaluated in Table\ \ref{tab:wb_results}. We find overestimates of AP of up to 6\% on BBBC010 with widely used AP implementations.

To avoid the potential pitfall of unintended non-comparability of AP values, AP is not recommended as a metric to perform comparative evaluations of proposal-free instance segmentation methods~\cite{maier_hein_2022_metrics_reloaded}. Instead, the kaggle data science bowl score S we also evaluate in Table\ \ref{tab:wb_results} is not based on confidence scores and is thus not affected by this pitfall.

\begin{table*}[h]
	\centering
	\caption{Impact of different / flawed AP implementations on BBBC010 results. 
	}
	\label{tab:wb_suppl_results}
	\begin{tabu} to 0.95\linewidth{ X[2.5l] X[2.0c] X[0.9c] X[0.9c] X[1.1c] X[1.1c] X[0.9c]}
		\toprule
		\multicolumn{7}{c}{BBBC010}\\
                AP implementation & avAP$_{[0.5:0.05:0.95]}$ & AP$_{0.5}$ & AP$_{0.6}$ & AP$_{0.7}$ & AP$_{0.8}$ & AP$_{0.9}$\\
                \midrule
                official COCO impl.  & 0.701 & 0.923 & 0.898 & 0.861 & 0.773 & 0.318\\
                official Cityscapes impl.  & 0.754 & 0.955 & 0.932 & 0.906 & 0.829 & 0.415\\
                fixed COCO impl. & 0.690 & 0.916 & 0.893 & 0.858 & 0.757 & 0.282\\
                fixed Cityscapes impl. & 0.694 & 0.923 & 0.893 & 0.859 & 0.761 & 0.283\\
                our impl. (p\(\cdot\)r) & 0.694 & 0.922 & 0.893 & 0.859 & 0.761 & 0.283\\
       \end{tabu}
     \end{table*}

\clearpage
\subsection{Supplemental List of Train/Val/Test Splits for BBBC010}
\label{suppl:list_split_bbbbc020}

\begin{table*}[h]
	\centering
	\caption{Train/val/test splits we employed to obtain the BBBC010 results reported in Table\ \ref{tab:wb_results}. We list respective individual image names from BBBC010. We recommend to use the split from \cite{chen_2022_layering} (bottom Table) as a future standard as it complies with the common practice in computer vision to not perform cross validation. }
	\label{tab:train_val_test_split}
	\begin{tabu} to 0.95\linewidth{ X[2.5l] X[10.0l] }
		\toprule
		\multicolumn{2}{c}{BBBC010}\\
                \midrule
                \multicolumn{2}{l}{50:50 split used for Table\ \ref{tab:wb_results} top, bottom, and middle-left:}\\
                \midrule
                training & A01 A02 A03 A04 A05 A06 A07 A08 A09 A10 A11 A12 A13 A14 A15 A16 A17 A18 A19 A20 A21 A22 A23 A24 B01 B02 B03 B04 B05 B06 B07 B08 B09 B10 B11 B12 B13 B14 B15 B16 B17 B18 B19 B20 B21 B22 B23 B24 C01 C02 \\
                \noalign{\vskip 0.5em}
                cross validation fold 1 & C03, C06, C07, C12, C13, C14, C15, C19, C21, C22, C24, D02, D03, D04, D05, D08, D10, D12, D14, D17, D18, D20, D22, D24, E03\\
                \noalign{\vskip 0.5em}
                cross validation fold 2 & C04, C05, C08, C09, C10, C11, C16, C17, C18, C20, C23, D01, D06, D07, D09, D11, D13, D15, D16, D19, D21, D23, E01, E02, E04\\
                \midrule
                \midrule
                \multicolumn{2}{l}{Split used for Table\ \ref{tab:wb_results} middle-right, following~\cite{chen_2022_layering}:}\\
                \midrule
                training & A01 A02 A03 A04 A05 A06 A07 A08 A09 A10 A11 A12 A13 A14 A15 A16 A19 A20 A21 A23 A24 B01 B02 B03 B04 B05 B06 B07 B08 B09 B10 B11 B12 B13 B14 B15 B16 B17 B18 B19 B20 B21 B22 B23 B24 C01 C03 C05 C06 C07 C08 C09 C10 C12 C13 C14 C15 C16 C18 C19 C20 C22 C23 C24 E01 E02 E03 E04\\
                \noalign{\vskip 0.5em}
                validation & A17 A18 A22 C02 C04 C11 C17 C21\\
                \noalign{\vskip 0.3em}
                testing & D01 D02 D03 D04 D05 D06 D07 D08 D09 D10 D11 D12 D13 D14 D15 D16 D17 D18 D19 D20 D21 D22 D23 D24\\
       \end{tabu}
     \end{table*}


\twocolumn
}{}

\end{document}